\documentclass[runningheads]{llncs}
\usepackage{graphicx}
\usepackage{comment}
\usepackage{amsmath,amssymb} 
\usepackage{color}

\usepackage{soul} 
\usepackage[dvipsnames]{xcolor}
\usepackage{booktabs}
\usepackage{multirow}
\usepackage{caption}
\usepackage{wrapfig}

\usepackage{subfig}
\usepackage{mysymbols}
\usepackage{paralist}
\usepackage{microtype}
\usepackage{arydshln}
\usepackage{chngcntr}
\usepackage{url}

\usepackage[pagebackref=true,breaklinks=true,letterpaper=true,colorlinks,bookmarks=false]{hyperref}

\newcommand{\datasetName}{TextCaps\xspace}
\newcommand{\datasetNCaptions}{142,040\xspace}
\newcommand{\datasetNCaptionsInK}{142k\xspace}
\newcommand{\datasetNImages}{28,408\xspace}
\newcommand{\myparagraph}[1]{\noindent\textbf{#1}}

\newcommand{\huapproach}{M4C\xspace}
\newcommand{\huapproachCaptioner}{\huapproach-Captioner\xspace}
\newcommand{\taskName}{image captioning with reading comprehension\xspace}
\newcommand{\TaskName}{Image captioning with reading comprehension\xspace}

\newcommand{\figcapmargin}{-1.0em}

\definecolor{demphcolor}{RGB}{124,124,124}
\newcommand{\demph}[1]{\textcolor{demphcolor}{#1}}
\newcommand{\icon}[1]{\includegraphics[height=10pt]{#1}}

\begin{document}
\pagestyle{headings}
\mainmatter

\title{\datasetName: a Dataset for Image Captioning\\with Reading Comprehension}

\titlerunning{\datasetName: a Dataset for Image Captioning with Reading Comprehension}

\author{Oleksii Sidorov\inst{1} \and
Ronghang Hu\inst{1,2} \and
Marcus Rohrbach\inst{1} \and
Amanpreet Singh\inst{1}}

\authorrunning{O. Sidorov et al.}

\institute{$^1$~Facebook AI Research $\qquad$ $^2$~University of California, Berkeley \\
\email{\{oleksiis,mrf,asg\}@fb.com}, \email{ronghang@eecs.berkeley.edu}}

\maketitle

\vspace{-1.5em}
\begin{abstract}

Image descriptions can help visually impaired people to quickly understand the image content.
While we made significant progress in automatically describing images and optical character recognition, current approaches are unable to include written text in their descriptions, although text is omnipresent in human environments and frequently critical to understand our surroundings. To study how to comprehend text in the context of an image we collect a novel dataset, TextCaps, with 145k captions for 28k images. Our dataset challenges a model to recognize text, relate it to its visual context, and decide what part of the text to copy or paraphrase, requiring spatial, semantic, and visual reasoning between multiple text tokens and visual entities, such as objects. We study baselines and adapt existing approaches to this new task, which we refer to as \emph{image captioning with reading comprehension}. Our analysis with automatic and human studies shows that our new TextCaps dataset provides many new technical challenges over previous datasets.

\end{abstract}

\section{Introduction}

\begin{figure}[t]
\centering
\includegraphics[width=.9\linewidth]{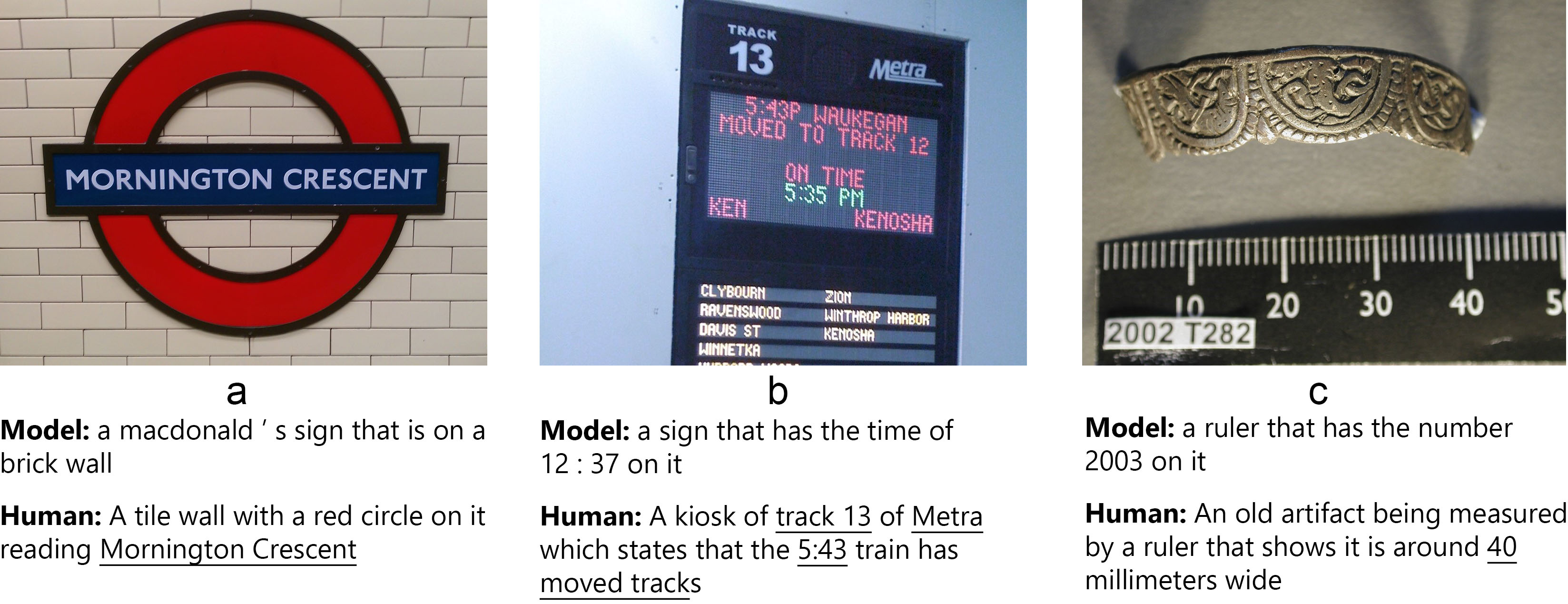}
\caption{\textbf{Existing captioning models cannot read!} \\ The \emph{\taskName} task using data from our TextCaps dataset and BUTD model \cite{butd} trained on it.} 

\label{fig:intro}
\end{figure}

When trying to understand man-made environments, it is not only important to recognize objects but also frequently critical to read associated text and comprehend it in the context to the visual scene. Knowing there is ``a red sign'' is not sufficient to understand that one is at ``Mornington Crescent'' Station (see Fig.~\ref{fig:intro}(a)), or knowing that an old artifact is next to a ruler is not enough to know that it is ``around 40 mm wide'' (Fig.~\ref{fig:intro}(c)). Reading comprehension in images is crucial for blind people. As the VizWiz datasets \cite{bigham2010vizwiz} suggest,  21\% of questions visually-impaired people asked about an image were related to the text in it. 
 Image captioning plays an  important role in starting a visual dialog with a blind user allowing them to ask for further information as required.
In addition, text out of context (\textit{e.g.} \myquote{5:43p}) may be of little help, whereas scene description (\textit{e.g.} `shown on a departure tableau') makes it substantially more meaningful. 

In recent years, with the availability of large labelled corpora, progress in image captioning has seen steady increase in performance and quality \cite{butd,chen2019uniter,Devlin2019BERTPO,Goyal2019ScalingAB,Wang2018GLUEAM} and reading scene text (OCR) has matured \cite{rosetta,he2018end,li2017towards,liu2018fots,smith2007overview}.
However, while OCR only focuses on written text, state-of-the-art image captioning methods focus only on the visual objects when generating captions and fail to recognize and reason about the text in the scene. For example, Fig.~\ref{fig:intro} shows predictions of a state-of-the-art model \cite{butd} on a few images that require reading comprehension. The predictions clearly show an inability of current state-of-the-art image captioning methods to read and comprehend text present in images. 
Incorporating OCR tokens into a sentence is a challenging task, as unlike conventional vocabulary tokens which depend on the text before them and therefore can be inferred, OCR tokens often can not be predicted from the context and therefore represent independent entities. Predicting a token from vocabulary and selecting an OCR token from the scene are two rather different tasks which have to be seamlessly combined to tackle this task.

Considering the images and reference captions in Fig.~\ref{fig:intro}, we can breakdown what is needed to successfully describe these images:First, detect and extract text/OCR tokens\footnote{The remainder of the manuscript we refer to the text in an image as ``OCR tokens'', where one token is typically a word, i.e. a group of characters.} (\myquote{Mornington Crescent}, \myquote{moved track}) as well the visual context such as objects in the image (\myquote{red circle}, \myquote{kiosk}). Second, generate a grammatically correct sentence which combines words from the vocabulary and OCR tokens. In addition to the challenges in normal captioning, \emph{\taskName}  can include the following technical challenges: 
 \begin{compactenum}
\item Determine the relationships \textbf{between different OCR tokens} and  between \textbf{OCR tokens and the visual context}, to decide if an OCR token should be mentioned in the sentence and which OCR tokens should be joined together (\eg in Fig. \ref{fig:intro}b: ``5:35'' denotes the current time and should not be joined with ``ON TIME''), based on their (a) \emph{semantics} (Fig. \ref{fig:preview}b), (b) \emph{spatial} relationship (Fig. \ref{fig:intro}c), and (c) \emph{visual} appearance and context (Fig. \ref{fig:preview}d).
\item  \textbf{Switching multiple times} during caption generation between the words from the model's vocabulary and OCR tokens (Fig. \ref{fig:intro}b).
\item  \textbf{Paraphrasing and inference} about the OCR tokens (Fig. \ref{fig:preview} bold).
\item  Handling of OCR tokens, including ones never seen before (\textbf{zero-shot}).
\end{compactenum}

While this list should not suggest a temporal processing order, it explains why today's models lack capabilities to comprehend text in images to generate meaningful descriptions. It is unlikely that the above skills will naturally emerge through supervised deep learning on existing image captioning datasets as they are not focusing on this problem. In contrast, captions in these datasets are collected in a way that implicitly or explicitly avoids mentioning specific instances appearing in the OCR text. 
To study the novel task of \taskName, we thus believe it is important to build a dataset containing captions which require reading and reasoning about text in images.
 
We find the COCO Captioning dataset \cite{coco}  not suitable  as only an estimated 2.7\% of its captions mention OCR tokens present in the image, and in total there are less than 350 different OCRs (i.e. the OCR vocabulary size), moreover most OCR tokens are common words, such as ``stop'', ``man'', which are already present in a standard captioning vocabulary.
Meanwhile, in Visual Question Answering, multiple datasets \cite{stvqa,ocrvqa,textvqa} were recently introduced which focus on text-based visual question answering. This task is harder than OCR recognition and extraction as it requires understanding the OCR extracted text in the context of the question and the image to deduce the correct answer. However, although these datasets focus on text reading, the  answers  are typically shorter than 5 words (mainly 1 or 2), and, typically, all the words which have to be generated  are either entirely from the training vocabulary \emph{or} OCR text, rather than requiring switching between them to build a complete sentence. 
These differences in task and  dataset do not allow training models to generate long sentences. Furthermore and importantly, we require a dataset with human collected reference sentences to validate and test captioning models for \emph{reading comprehension}.

Consequently,  in this work, we contribute the following:
\begin{compactitem}
    \item For our novel task \emph{\taskName}, we collect a new dataset, \textbf{\datasetName}, which \textbf{contains \datasetNCaptions captions} on \datasetNImages images and requires models to read and reason about text in the image to generate coherent descriptions.
    \item We analyse our dataset, and find it has \textbf{several new technical challenges for captioning}, including the ability to  switch multiple times between OCR tokens and vocabulary, zero-shot OCR tokens, as well as paraphrasing and inference about OCR tokens.
    \item Our evaluation shows  that \textbf{standard captioning models fail on this new task}, while the 
    state-of-the-art TextVQA \cite{textvqa} model, M4C \cite{hu}, when trained with our dataset \datasetName, gets encouraging results. Our ablation study shows that it is important to take into account all semantic, visual, and spatial information of OCR tokens to generate high-quality captions. 
    \item We conduct \textbf{human evaluations} on model predictions which show that there is a \textbf{significant gap between the best model and humans}, indicating an exciting avenue of future image captioning research. 
\end{compactitem}

\section{Related work}

\myparagraph{Image Captioning.}
The Flickr30k \cite{flickr30k} and COCO Captions \cite{coco} dataset have both been collected similarly via crowd-sourcing.The COCO Captions dataset is significantly larger than Flickr30k and acts as a base for training the majority of current state-of-the-art image captioning algorithms. It includes 995,684 captions for 164,062 images. The annotators of COCO were asked ``Describe all the important parts of the scene" and ``Do not describe unimportant details", which resulted in COCO being focused on objects which are more prominent rather than text. 
SBU Captions~\cite{sbu} is an image captioning dataset which was collected automatically by retrieving one million images and associated user descriptions from Flickr, filtering them  based on key words and sentence length.
Similarly, Conceptual Captions (CC) dataset \cite{cc} is also automatically constructed by crawling images from web pages together with their ALT-text. 
The collected annotations were extensively filtered and processed, e.g. replacing proper names and titles with object classes (\eg man, city), resulting in 3.3 million image-caption pairs.
This simplifies caption generation but at the same time removes fine details such as unique OCR tokens. Apart from conventional paired datasets there are also datasets like NoCaps~\cite{nocaps}, oriented to a more advanced task of captioning with zero-shot generalization to novel object classes.

While our \datasetName dataset also consists of image-sentence pairs, it focuses on the text in the image, posing additional challenges. Specifically, text can be seen as an additional modality, which models  have to read (typically using OCR), comprehend, and include when generating a sentence. Additionally, many OCR tokens do not appear in the training set , but only in the test (zero-shot). In concurrent work, \cite{gurari2020captioning} collect captions on VizWiz \cite{bigham2010vizwiz} images but unlike TextCaps there isn't a specific focus on reading comprehension.

\myparagraph{Optical Character Recognition (OCR).}
OCR involves in general two steps, namely (i) detection: finding the location of text, and (ii) extraction: based on the detected text boundaries, extracting the text as characters. OCR can be seen as a subtask for our \emph{\taskName} task as one needs to know the text present in the image to generate a meaningful description of an image containing text. This makes OCR research an important and relevant topic to our task, which additionally requires to understand the importance of OCR token, their semantic meaning, as well as relationship to visual context and other OCR tokens. 
Recent OCR models have shown reliability and performance improvements \cite{rosetta,smith2007overview,li2017towards,liu2018fots,he2018end}. However, in our experiments we observe that OCR is far from a solved problem in real-world scenarios present in our dataset.

\myparagraph{Visual Question Answering with Text Reading Ability.}
Recently, three different text-oriented datasets were presented for the task of Visual Question Answering. TextVQA \cite{textvqa} consists of 28,408 images from selected categories of Open Images v3 dataset, corresponding 45,336 questions, and 10 answers for each question. Scene Text VQA (ST-VQA) dataset \cite{stvqa} has a similar size of 23,038 images and 31,791 questions but only one answer for each question. Both these datasets were annotated via crowd-sourcing. OCR-VQA \cite{ocrvqa} is a larger dataset (207,572 images) collected semi-automatically using photos of book covers and corresponding metadata. 
The rule generated questions were paraphrased by human annotators.
These three datasets require reading and reasoning about the text in the image while considering the context for answering a question, which is similar in spirit to \datasetName. However, the image, question and answer triplet is not directly suitable for generation of descriptive sentences. We provide additional quantitative comparisons and discussion between our and existing captioning and VQA datasets in Section \ref{data_analysis}.
\section{ \protect\icon{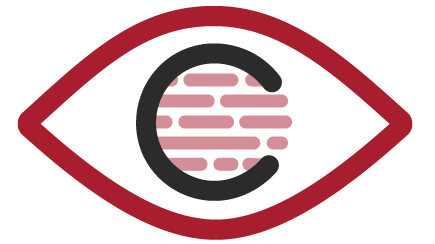} \datasetName Dataset }

We collect \datasetName with the goal of studying the novel task of \emph{\taskName}. Our dataset allows us to test captioning models' reading comprehension ability and we hope it will also enable us to teach image captioning models how ``to read", \emph{i.e.}, allow us to design and train image captioning algorithms which are able to process and include information from the text in the image. 
In this  section, we describe the dataset collection and analyze its statistics. 
The dataset is publicly available at \href{https://textvqa.org/textcaps}{textvqa.org/textcaps}.

\subsection{Dataset collection}
\begin{figure*}[t]
\centering
\includegraphics[width=.9\linewidth]{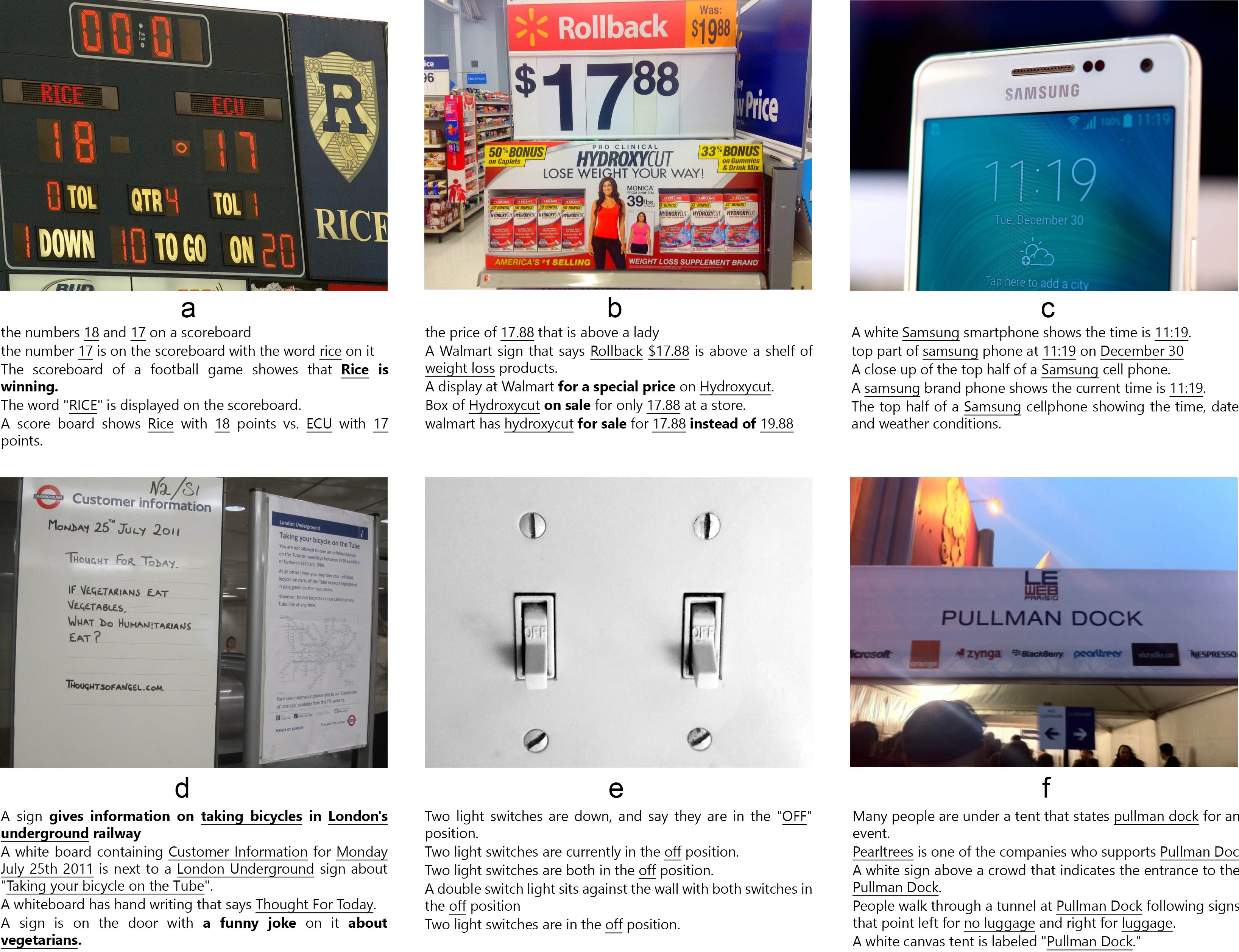}
\caption{\textbf{Illustration of \datasetName captions.} The bold font highlights instances which do not copy the text directly but require  paraphrasing or some inference beyond copying. Underlined font highlights copied text tokens.} 
\label{fig:preview}
\end{figure*}

With the goal of having a diverse set of images, we rely on images from Open Images v3 dataset (CC 2.0 license). 
Specifically, we use the same subset of images as in the TextVQA dataset \cite{textvqa}; these images have been verified to contain text through an OCR system \cite{rosetta} and human annotators \cite{textvqa}. Using the same images as TextVQA additionally allows multi-task and transfer learning scenarios between OCR-based VQA and image captioning tasks. 
The images were annotated by human annotators in two stages.\footnote{The full text of the instructions as well as screenshots of the user interface are presented in the Supplemental (Sec.~\ref{sec:user-interfache}).}
\begin{compactitem}
\item[\textbf{Annotators}] were asked to describe an image in one sentence which would require reading the text in the image.\footnote{Apart from direct copying, we also allowed indirect use of text, \emph{e.g.} inferring, paraphrasing, summarizing, or reasoning about it (see Fig. \ref{fig:preview}). This approach creates a fundamental difference from OCR datasets where alteration of text is not acceptable. For captioning, however, the ability to reason about text can be beneficial.}
\item[\textbf{Evaluators}] were asked to vote yes/no on whether the caption written in the first step satisfies the following requirements: requires reading the text in the image; is true for the given image; consists of one sentence; is grammatically correct; and does not contain subjective language.
The majority of 5 votes was used to filter captions of low quality. The quality of the work of evaluators was controlled using gold captions of known good/bad quality. 
\end{compactitem}
Five independent captions were collected for each image. An additional 6th caption was collected for the test set only to estimate human performance on the dataset.
The annotators did not see previously collected captions for a particular image and did not see the same image twice.
In total, we collected 145,329 captions for 28,408 images. We follow the same image splits as TextVQA for training (21,953), validation (3,166), and test (3,289) sets. An estimation performed using ground-truth OCR shows that on average, 39.5\% out of all OCR tokens present in the image are covered by the collected human annotations. 

\subsection{Dataset analysis}
\label{data_analysis}

We first discuss several properties of the \datasetName qualitatively and then analyse and compare its statistics to other  captioning and OCR-based VQA datasets.

\myparagraph{Qualitative observations.}
Examples of our collected dataset  in Fig.~\ref{fig:preview} demonstrate that our image captions combine the textual information present in the image with its natural language scene description. We asked the annotators to read and use text in the images but we did not restrict them to directly copy the text. Thus, our dataset also contains captions where OCR tokens are not present directly but were used to infer a description, \emph{e.g.} in Fig. \ref{fig:preview}a ``Rice is winning" instead of ``Rice has 18 and Ecu has 17". 
In a human evaluation of 640 captions we found that about 20$\%$ of images have at least one caption (8\% of captions) which require more challenging reasoning or paraphrasing rather than just direct copying of visible text.
Nevertheless, even the captions which require copying text directly can be complex and may require advanced reasoning as illustrated in multiple examples in Fig.~\ref{fig:preview}. The collected captions are not limited to trivial template ``Object \textit{X} which says \textit{Y}". We have observed various types of relations between text and other objects in a scene which are impossible to formulate without reading comprehension. For example, in Fig.~\ref{fig:preview}: 
``A \textit{score board} shows \underline{Rice} with \underline{18} points vs. \underline{ECU} with \underline{17} points" (a), ``\textit{Box} of \underline{Hydroxycut} on sale for only \underline{17.88} at a \textit{store}'' (b),  ``Two \textit{light switches} are both in \underline{off} position" (e).  

\begin{figure}[t!]
\centering
\makebox[\textwidth][c]{\includegraphics[width=1.1\linewidth]{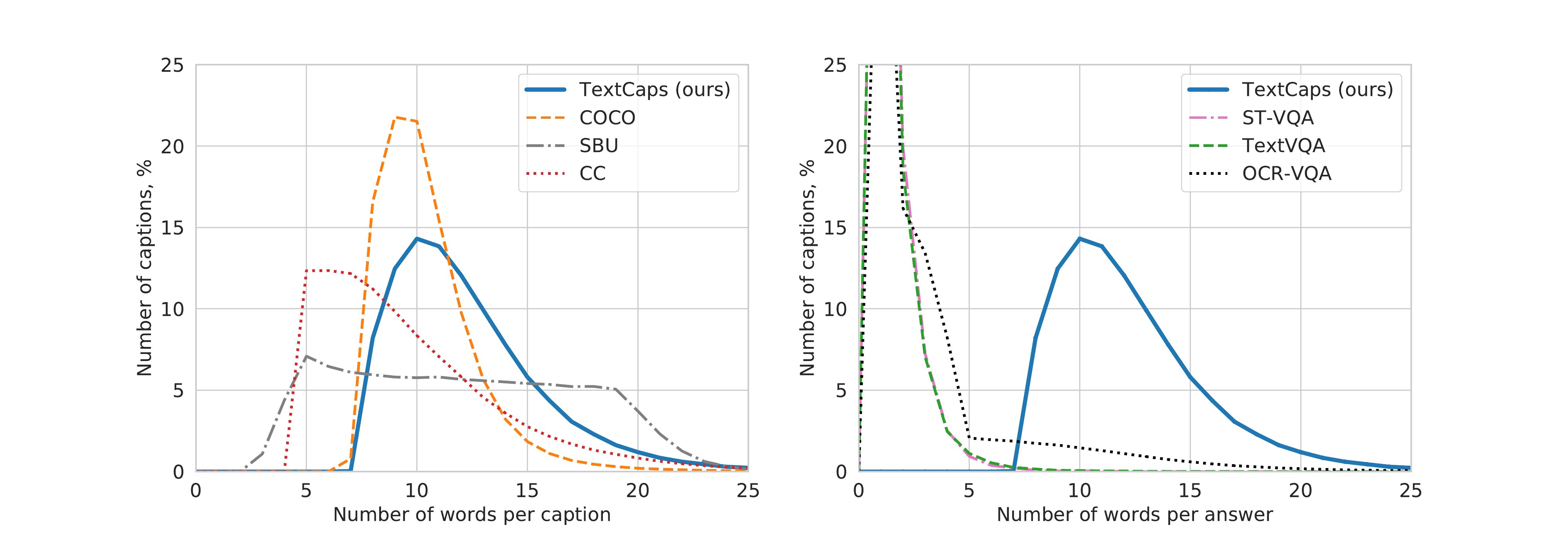}}%

\caption{\textbf{Distribution of caption/answer lengths} in Image Captioning (left) and VQA (right) datasets. VQA answers are significantly shorter than image captions and mostly concentrated within 5 words limit.}
\label{fig:length}
\end{figure}

\myparagraph{Dataset statistics.}
To situate \datasetName properly \wrt other image captioning datasets, we compare \datasetName with other prominent image captioning datasets, namely COCO \cite{coco}, SBU \cite{sbu}, and Conceptual Captions \cite{cc}, as well as reading-oriented VQA datasets TextVQA \cite{textvqa}, ST-VQA \cite{stvqa}, and OCR-VQA \cite{ocrvqa}.

The average caption length is 12.0 words for SBU, 9.7 words for Conceptual Captions, and 10.5 words for COCO, respectively. The average length for \datasetName  is 12.4, slightly larger than the others (see Fig. \ref{fig:length}). This can be explained by the fact that captions in TextCaps typically include both scene description as well as the text from it in one sentence, while conventional captioning datasets only cover the scene description. Meanwhile, the average answer length is 1.53 for TextVQA, 1.51 for ST-VQA and 3.31 for OCR-VQA -- much smaller than the captions in our dataset. Typical answers like \textit{`yes'}, \textit{`two'}, \textit{`coca cola'} may be sufficient to answer a question but insufficient to describe the image comprehensively.

\begin{figure}[t!]
\centering
\includegraphics[width=1\linewidth]{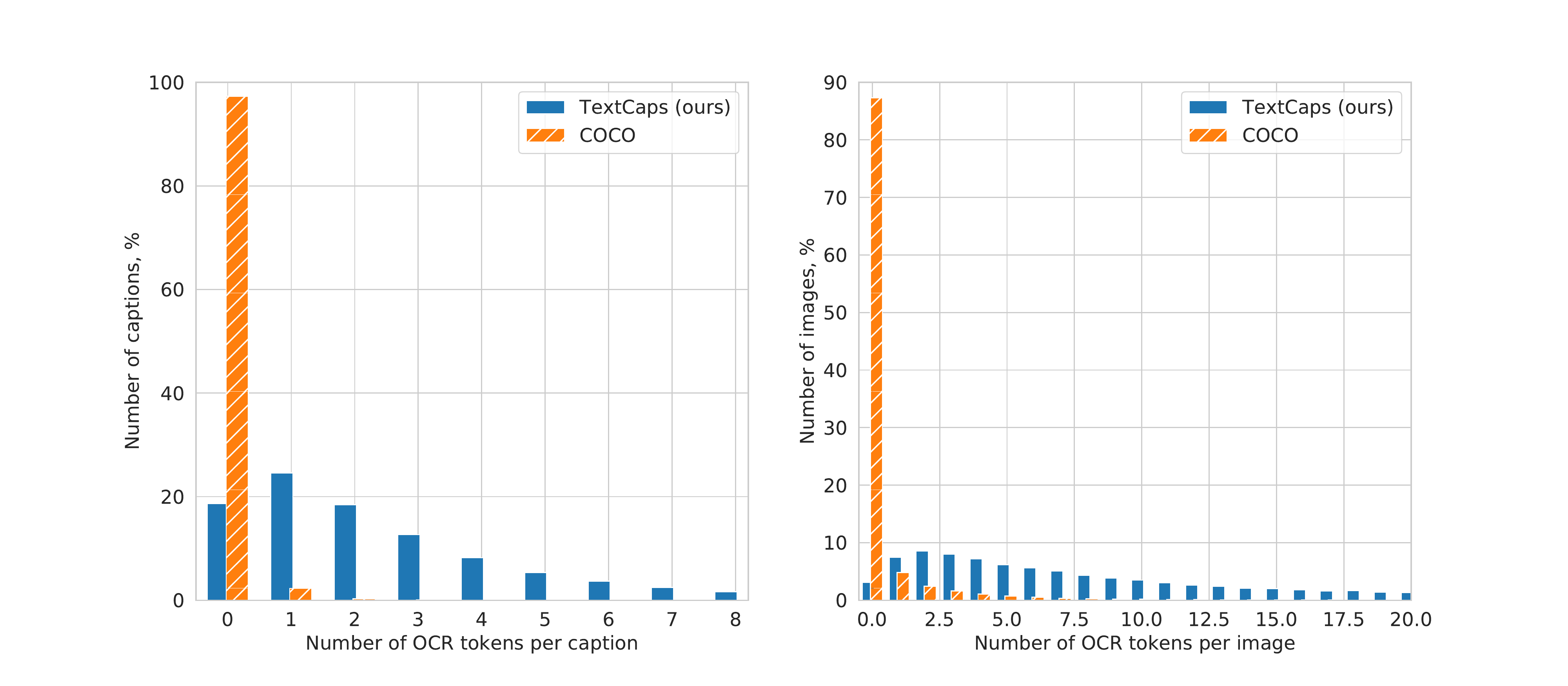}
\caption{\textbf{Distribution of OCR tokens} in COCO and TextCaps captions (left) and images (right). In total, COCO contains \textbf{2.7\%} of captions and \textbf{12.7\%} of images with at least one OCR token, whereas \datasetName\  -- \textbf{81.3\%} and \textbf{96.9\%}. }
\label{fig:ocr_images}

\subfloat[\textbf{OCR frequency distribution}  shows how many OCR tokens occur once, twice, \textit{etc.} TextCaps has the largest amount of unique and rare ($<5$) OCR tokens. Note that TextVQA has 10 answers for each question which are often identical.\label{fig:frequency}]{
\includegraphics[width=0.48\linewidth]{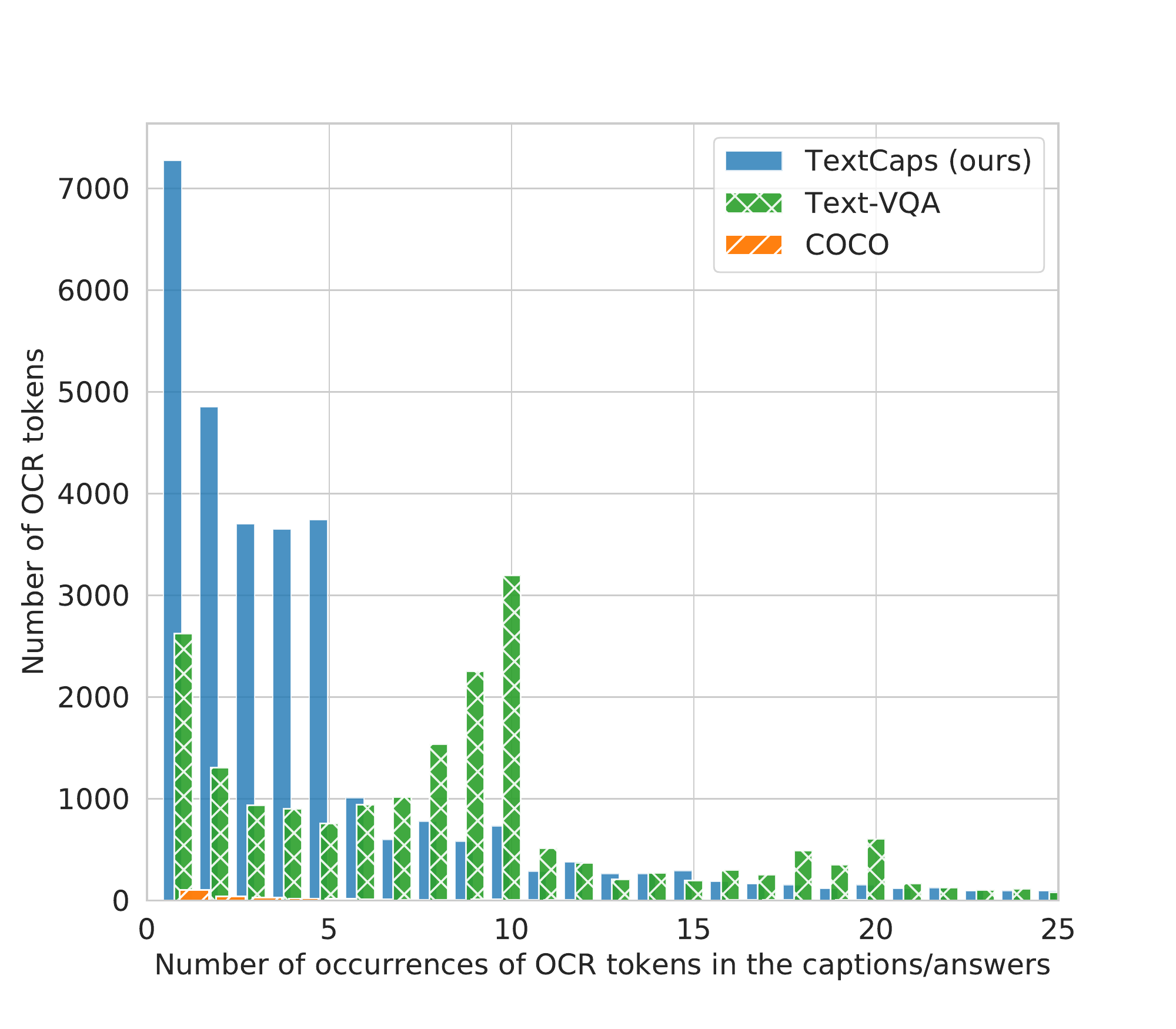}}\hspace{.02\linewidth}
\subfloat[\textbf{Number of switches between OCR $\rightleftharpoons$ Vocab} illustrates the technical complexity of the datasets. An approach which cannot make switches will be sufficient for most of COCO captions and TextVQA but not for TextCaps.\label{fig:switches}]{ \includegraphics[width=0.47\linewidth]{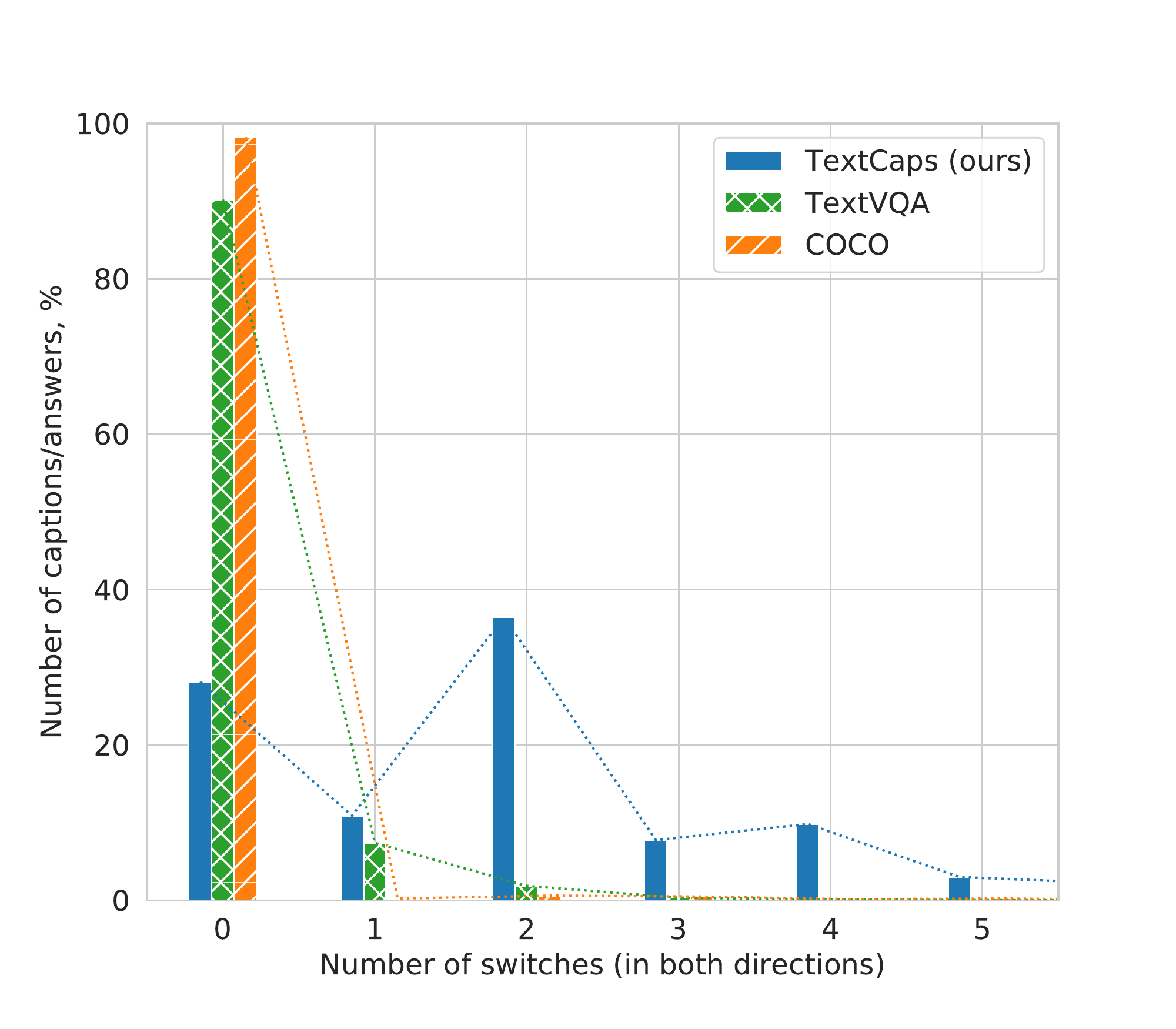}}
\caption{Analysis of OCR in our dataset vs. others}
\end{figure}

Fig.~\ref{fig:ocr_images} compares the percentage of captions with a particular number of OCR tokens between COCO and \datasetName datasets.\footnote{Note that OCR tokens are extracted using Rosetta OCR system \cite{rosetta} which cannot guarantee exhaustive coverage of all text in an image and presents just an estimation.} \datasetName has a much larger number of OCR tokens in the captions as well as in the images compared to COCO (note the high percentage at 0). A small part (2.7\%) of COCO captions which contain OCR tokens is mostly limited to one token per caption; only 0.38\% of captions contain two or more tokens. Whereas in \datasetName, multi-word reading is much more common (56.8\%) which is crucial for capturing real-world information (\eg authors, titles, monuments, \textit{etc}.). Moreover, while COCO Captions contain less than 350 unique OCR tokens, TextCaps contains 39.7k of them.

We also measured the  frequency of OCR tokens  in the captions. Fig.~\ref{fig:frequency} illustrates the number of times a particular OCR token appears in the captions. More than 9000 tokens appear only once in the whole dataset. The curve drops rapidly after 5 occurrences and only a small part of tokens occur more than 10 times. Quantitatively, 75.7\% of tokens are presented less then 5 times, and only 12.9\% are presented more than 10 times. The distribution specifically demonstrates the large variance in text occurring in natural images which is challenging to model using a fixed word vocabulary. 

In addition to this long-tailed distribution, we  find that an impressive number of 2901 of 6329 unique OCR tokens appearing in the test set captions, have neither appeared in the training nor validation set (i.e. they are ``zero-shot") which makes it necessary for models  to be able to read new text in images. 

TextCaps dataset also creates new technical challenges for the models. Figure~\ref{fig:switches} illustrates that due to the common use of OCR tokens in the captions, models required to switch between OCR and vocabulary words often. The majority of the TextCaps captions require to switch twice or more, whereas most COCO and TextVQA outputs can be generated even without any switches.

\section{Benchmark Evaluation}

\subsection{Baselines}
\label{sec:exp_baselines}

Our baselines aim to illustrate the gap between performance of conventional state-of-the-art image captioning models (BUTD \cite{butd}, AoANet\cite{aoa}) in comparison to recent architectures which incorporate reading (M4C \cite{hu}). 

\myparagraph{Bottom-Up Top-Down Attention model (BUTD)} \cite{butd} is a widely used image captioning model based on Faster R-CNN \cite{ren2015faster} object detection features (Bottom-Up) in conjunction with attention-weighted LSTM layers (Top-Down). 

\myparagraph{Attention on Attention model (AoANet)}
\cite{aoa} is a current SoTA captioning algorithm which uses the attention-on-attention module (AoA) to create a relation between attended vectors in both encoder and decoder.

\begin{figure*}[t]
\centering
\includegraphics[width=0.85\linewidth]{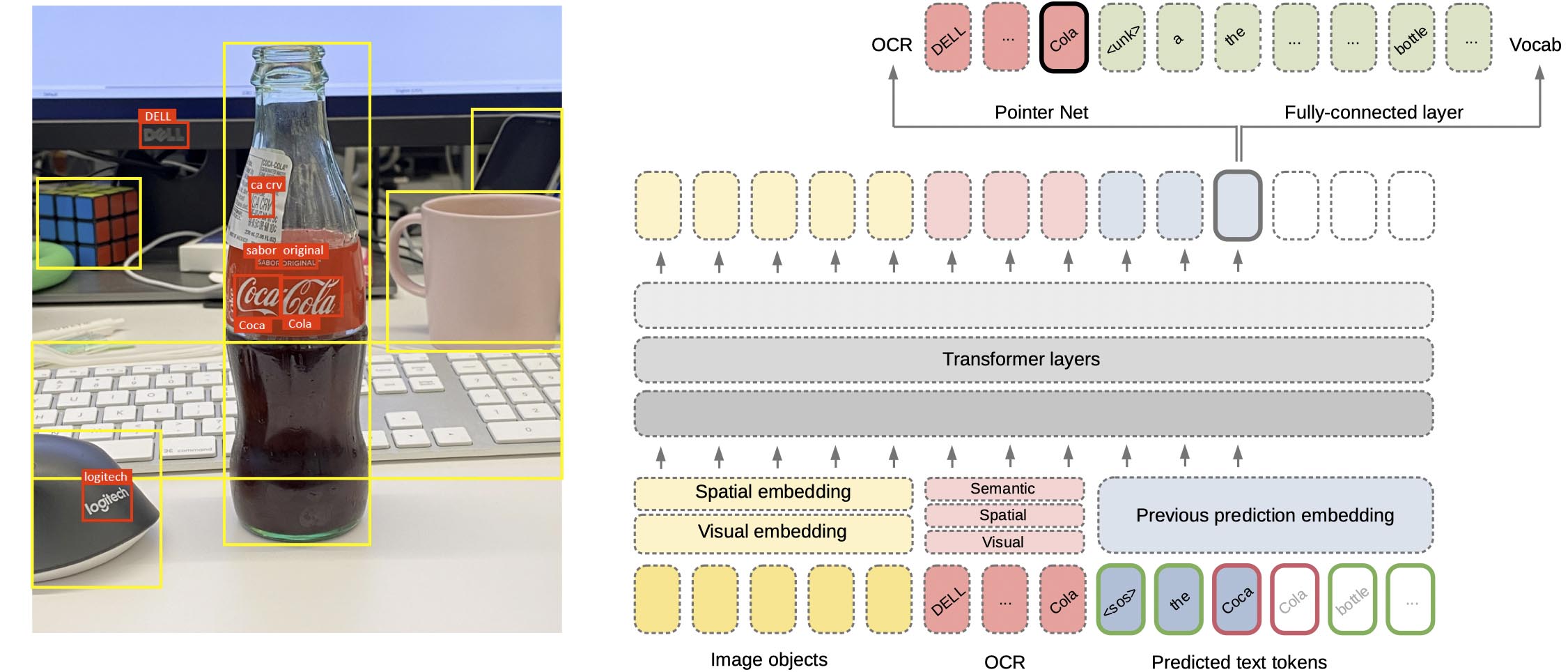}
\caption{\textbf{M4C-Captioner architecture} 
for the \taskName task. }
\label{fig:m4c}
\end{figure*}

\myparagraph{\huapproachCaptioner.}
\huapproach \cite{hu} is a recent model with state-of-the-art performance on the TextVQA task. The model fuses different modalities by embedding them into a common semantic space and processing them with a multimodal transformer. Apart from that, unlike conventional VQA models where a prediction is made via classification, it enables iterative answer decoding with a dynamic pointer network \cite{pointer2,pointer1}, allowing the model to generate a multi-word answer, which is not limited to a fixed vocabulary. This feature makes it also suitable for reading-based caption generation. We adapt \huapproach to our task by  removing the question input and directly use its multi-word answer decoder to generate a caption conditioned on the detected objects and OCR tokens in the image  (we refer to this model as \textbf{\huapproachCaptioner} and illustrate it in Figure \ref{fig:m4c}). 

\myparagraph{\huapproachCaptioner ablations.}
In comparison to its full version, we also evaluate a restricted version of this model without access to OCR results (referred to as \textbf{\huapproachCaptioner w/o OCRs}), where we use an empty OCR token list as input to the model. Additionally, we experiment with removing the pointer network (described in details in \cite{hu}) from \huapproachCaptioner, so that the model still has access to OCR features but cannot directly copy OCR tokens, and must use its fixed vocabulary for caption generation (referred to as \textbf{\huapproachCaptioner w/o copying}). 
As multiple types of features are used for OCR tokens in \huapproachCaptioner by default (same as in \cite{hu}), we further study the impact of each OCR feature type and use only \textbf{spatial} information (4-dimensional relative bounding box coordinates $[x_{\text{min}}, y_{\text{min}}, x_{\text{max}}, y_{\text{max}}]$ of OCR tokens), \textbf{semantic} information (FastText \cite{bojanowski2017enriching} and PHOC \cite{almazan2014word}), and \textbf{visual} (Faster R-CNN \cite{ren2015faster}) features in different experiments. Additionally, we use ground truth OCR tokens annotated by humans (referred to as \textbf{\huapproachCaptioner w/ GT OCRs}) for training and prediction\footnote{This includes a small number of images without GT-OCRs (Supplemental Sec.~\ref{sec:gt-ocr}).} to study the influence of mistakes of automatic OCR methods.

\myparagraph{Human performance.}
In addition to our baselines, we provide an estimate of human performance by using the same metrics on the \datasetName test set to benchmark the progress that models still need to make. 
As discussed in Section \ref{sec:exp_textcaps}, we collected one more caption for each image in the test set. 
The metrics are then calculated by averaging the results over 6 runs, each time leaving out one caption as a prediction, similar to \cite{vqametric}. On the test set, we use the same approach to evaluate machine-generated captions, so numbers are comparable. 

\subsection{Experimental setup}\footnote{Code for experiments is available at \href{https://git.io/JJGuG}{https://git.io/JJGuG}}
We follow the default configurations and hyper-parameters for training and evaluation of each baseline. For AoANet we use original implementation and feature extraction technique. For BUTD \cite{butd}, we use the implementation and hyper-parameters from MMF \cite{singh2018pythia,singh2020mmf}. For \huapproachCaptioner \cite{hu}, we follow the same implementation details as used for TextVQA task \cite{hu}. We train both models for the same number of iterations on the \datasetName training set. During caption generation, we remove the \texttt{<unk>} token (for unknown words).

\begin{table}[t]
\setlength{\tabcolsep}{3pt}
\scriptsize
\caption{\textbf{Performance of our baselines on our \datasetName dataset.} \huapproachCaptioner significantly benefits from OCR inputs and achieves the highest CIDEr score, suggesting that it is important to copy text from image on this task. However, there is still a large gap between the current machine performance and human performance, which we hope can be closed by future work.}
\begin{center}
\begin{tabular}{cp{5.3cm}ccccccc}
\cmidrule[\heavyrulewidth]{1-8}
& & & \multicolumn{5}{c}{\datasetName validation set metrics}  \\
\cmidrule{4-8} 
\# & Method & Trained on 
& B-4 & M & R & S & C \\
\cmidrule{1-8}
1 & BUTD \cite{butd} & COCO 
& 12.4 & 13.3 & 33.7 & 8.7 & 24.2 \\
2 & BUTD \cite{butd} & \datasetName 
& 20.1 & 17.8 & 42.9 & 11.7 & 41.9 \\
3 & AoANet \cite{aoa} & COCO 
& 18.1 & 17.7 & 41.4 & 11.2 & 32.3 \\
4 & AoANet \cite{aoa} & \datasetName
& 20.4 & 18.9 & 42.9 & 13.2 & 42.7 \\
5 & \huapproachCaptioner & COCO
& 12.3 & 14.2 & 34.8 & 9.2 & 30.3 \\
6 & \huapproachCaptioner & TextVQA
& 0.1 & 4.4 & 11.3 & 2.8 & 16.9 \\
7 & \huapproachCaptioner w/o OCRs & \datasetName 
& 15.9 & 18.0 & 39.6 & 12.1 & 35.1 \\
8 & \huapproachCaptioner w/o copying & \datasetName 
& 18.2 & 19.2 & 41.5 & 13.1 & 49.2 \\
9 & \huapproachCaptioner (OCR semantic) & \datasetName 
& 21.4 & 20.4 & 44.0 & 14.1 & 69.0 \\
10 & \huapproachCaptioner (OCR spatial) & \datasetName 
& 21.7 & 20.6 & 44.6 & 13.7 & 72.0 \\
11 & \huapproachCaptioner (OCR visual) & \datasetName 
& 22.5 & 21.3 & 45.3 & 14.4 & 84.0 \\
12 & \huapproachCaptioner (OCR semantic \& visual) & \datasetName 
& 23.4 & 21.5 & 45.8 & 14.9 & 86.0 \\
13 & \huapproachCaptioner & \datasetName 
& \textbf{23.3} & \textbf{22.0} & \textbf{46.2} & \textbf{15.6} & \textbf{89.6} \\ \cmidrule{1-8}
14 & \huapproachCaptioner (w/ GT OCRs)& \datasetName 
& \demph{26.0} & \demph{23.2} & \demph{47.8} & \demph{16.2} & \demph{104.3} \\
\cmidrule[\heavyrulewidth]{1-8} 
& & & \multicolumn{6}{c}{\datasetName test set metrics} \\
\cmidrule{4-9} 
\# & Method & Trained on 
& B-4 & M & R & S & C &  \multicolumn{1}{|c}{H}  \\
\cmidrule{1-9}
15 & BUTD \cite{butd} & \datasetName 
& 14.9 & 15.2 & 39.9 & 8.8 & 33.8 & \multicolumn{1}{|c}{1.4} \\
16 & AoANet \cite{aoa} & \datasetName
& 15.9 & 16.6 & 40.4 & 10.5 & 34.6 &  \multicolumn{1}{|c}{1.4}  \\
17 & \huapproachCaptioner & \datasetName 
& \textbf{18.9} & \textbf{19.8} & \textbf{43.2} & \textbf{12.8} & \textbf{81.0} &  \multicolumn{1}{|c}{\textbf{3.0}}\\ \cmidrule{1-9}
18 & \huapproachCaptioner (w/ GT OCRs)& \datasetName 
& \demph{21.3} & \demph{21.1} & \demph{45.0} & \demph{13.5} & \demph{97.2} &  \multicolumn{1}{|c}{\demph{3.4}}  \\
19 & Human & -- 
& 24.4 & 26.1 & 47.0 & 18.8 & 125.5 &  \multicolumn{1}{|c}{4.7} \\
\bottomrule
\end{tabular} \\
B-4: BLEU-4; M: METEOR; R: ROUGE\_L; S: SPICE; C: CIDEr; H: human evaluation
\end{center}
\label{tab:exp_textcaps_method_comparison}
\end{table}

\myparagraph{Datasets.} 
We first evaluate the models trained using COCO dataset on \datasetName to demonstrate how existing datasets and models lack reading comprehension. Then we train and evaluate each baseline using \datasetName.

\myparagraph{Metrics.} Apart from automatic captioning metrics including BLEU \cite{papineni2002bleu}, METEOR \cite{denkowski2014meteor}, ROUGE\_L \cite{lin2004rouge}, SPICE \cite{anderson2016spice}, and CIDEr \cite{vedantam2015cider}, we also perform human evaluation. 
We collect 5000 human scores on a Likert scale from 1 to 5 
\begin{wrapfigure}[15]{r}{0.4\textwidth}
\centering
\includegraphics[width=\linewidth,trim={0 0 0 2em},clip]{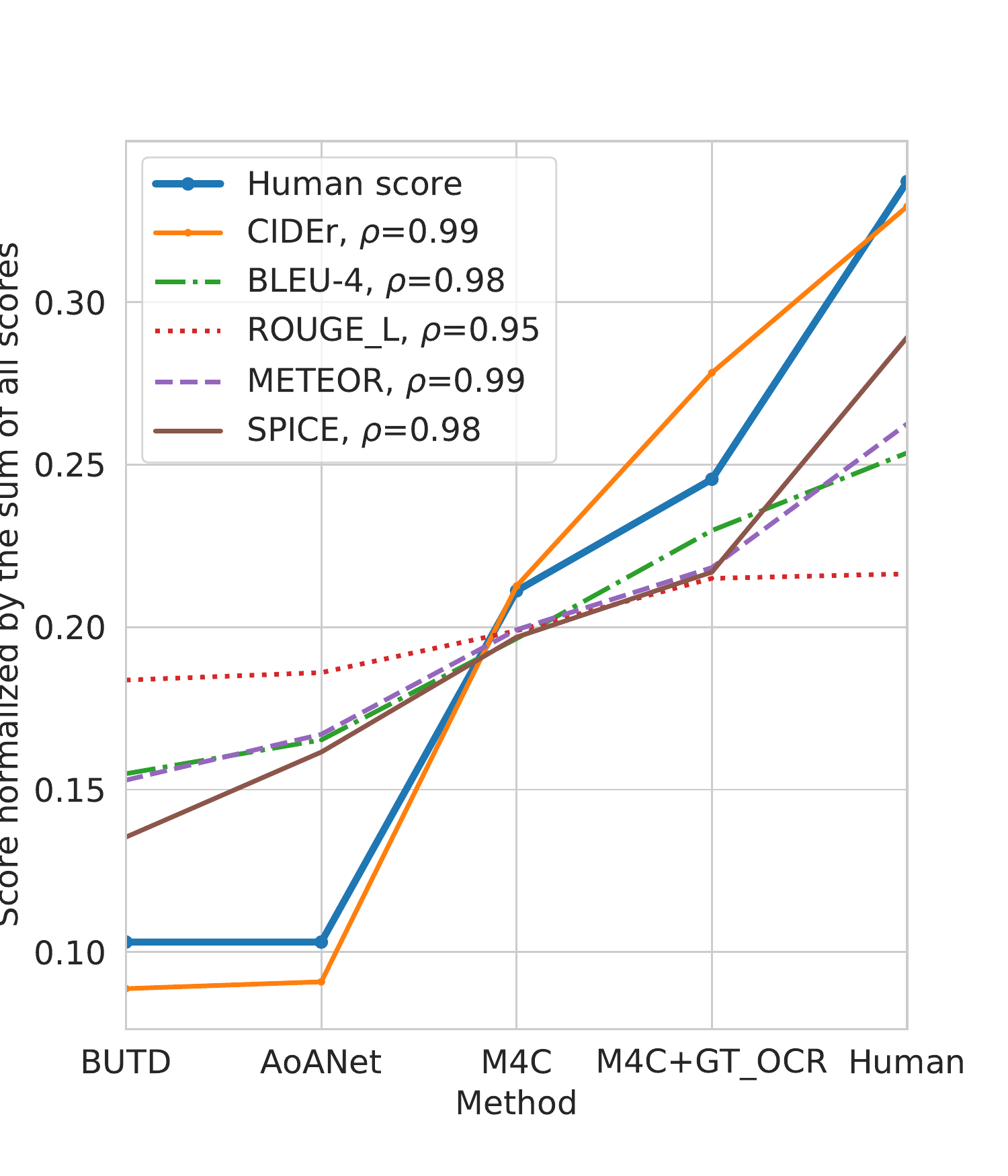}
\caption{\textbf{Human evaluation in comparison to automatic metrics.}}
\label{fig:metric}
\end{wrapfigure}
for a random sample of 200 images and compute  median score for each caption. 
Fig. \ref{fig:metric} shows that ranking of the sentence quality is the same as for automatic metrics. Moreover, all the metrics show very high correlation with human scores but CIDEr and METEOR have the highest. For comparison between different methods, we focus on the CIDEr, which puts more weight on informative n-grams in the captions (such as OCR tokens) and less weight on commonly occurring words with TF-IDF weighting.

\subsection{Results}
\subsubsection{\datasetName Dataset.}
\label{sec:exp_textcaps}

\begin{figure*}[t!]
\centering
\includegraphics[width=.9\linewidth]{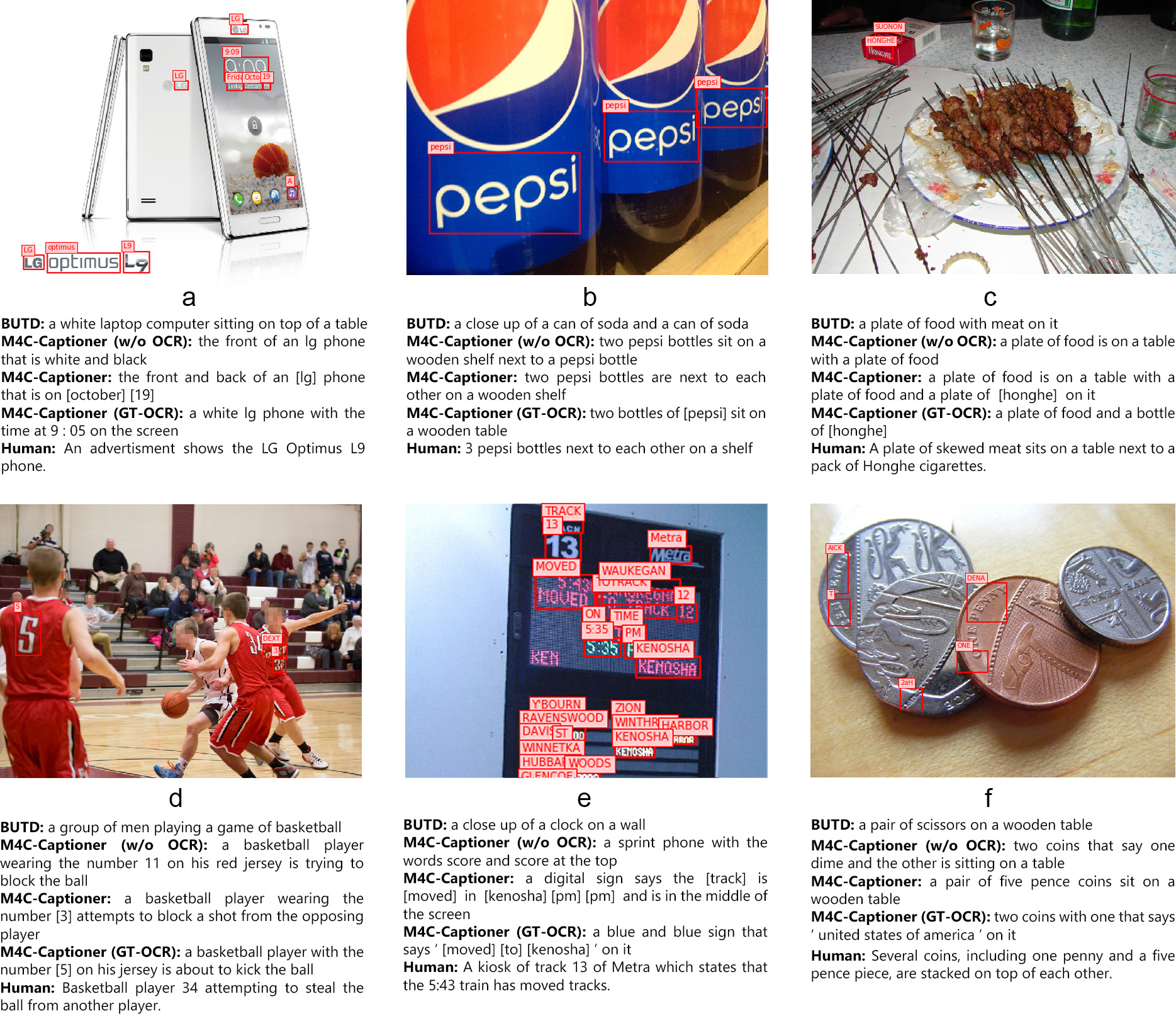}
\caption{\textbf{Illustration of positive and negative predictions from different models} on \datasetName validation set. For M4C-Captioner, square brackets indicate tokens copied from OCR. While most of the time OCR tokens are very important for correct copying of the text from the images, for common terms such as ``pepsi" or ``pence", the model sometimes prefer to select them from the vocabulary.}
\label{fig:m4c-pred}
\end{figure*}

It can be observed in results (Table~\ref{tab:exp_textcaps_method_comparison}) that the BUTD model trained on the COCO captioning dataset (line 1) achieves the lowest CIDEr score, indicating that it fails to describe text in the image. When trained on the \datasetName dataset (line 2), the BUTD model has higher scores as expected, since there is no longer a domain shift between training and evaluation. AoANet (line 3, 4), which is a stronger captioning model, outperforms BUTD but still cannot handle reading comprehension and largely underperforms M4C-Captioner. For the \huapproachCaptioner model, there is a large gap (especially in CIDEr scores) between training with and without OCR inputs (line 13 vs. 7).
Moreover, ``\huapproachCaptioner w/o copying'' (line 8) is worse than the full model (line 13) but better than the more restricted ``\huapproachCaptioner w/o OCRs'' (line 7). The results indicate that it is important to both encode OCR features \textbf{and} be able to directly copy OCR tokens.
We also observe (in line 13 vs. 9-12) that it is important for a model to use spatial, visual, and semantic features of OCR tokens together, especially in the complex combinations of OCR tokens where both spatial relation and semantics play an important role in finding a connection between words. However, on the test set, we still notice a large gap between the best machine performance (line 17) and the human performance (line 19) on this task. 
Also, using ground-truth OCRs (line 18) reduces this gap but still does not close it, suggesting that there is room for future improvement in both better reasoning and better text recognition.

\begin{figure*}[t]
\centering
\includegraphics[width=.9\linewidth]{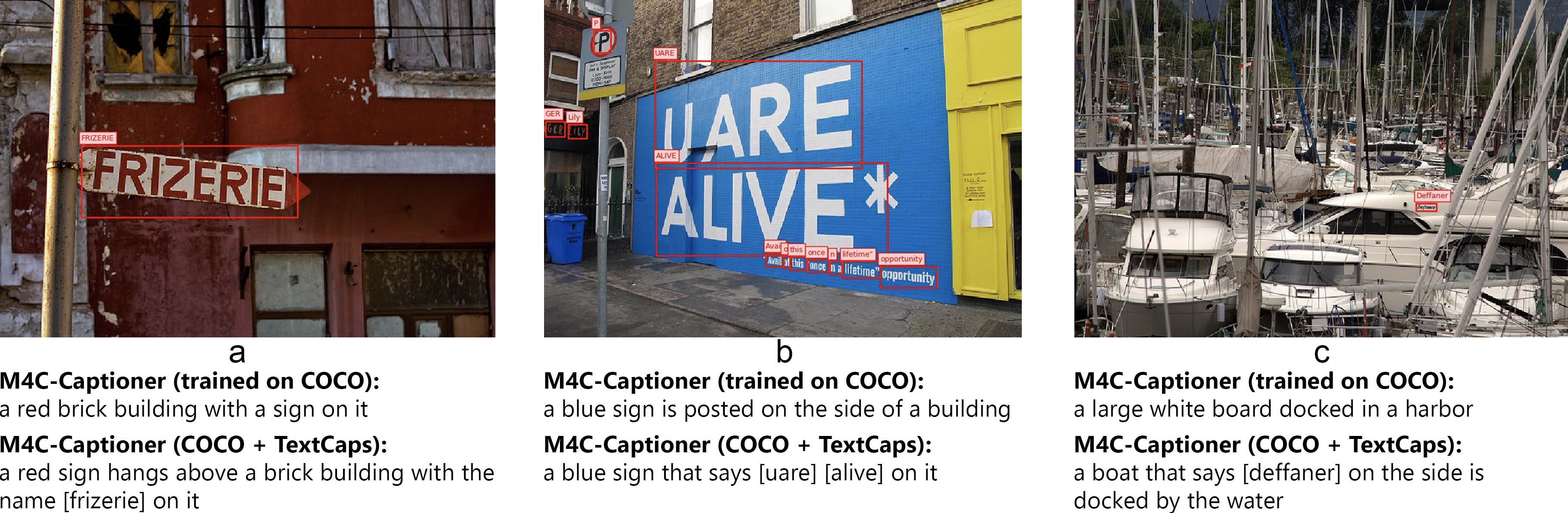}
\caption{\textbf{Examples of M4C-Captioner's predictions on COCO data} when trained on COCO and \datasetName. It can be observed that \textit{despite of availability of OCR module in both cases}, using \datasetName pushes model to read the text. Square brackets indicate tokens copied from OCR.}
\label{fig:m4c-pred-coco}
\end{figure*}

Figure~\ref{fig:m4c-pred} shows qualitative examples from different methods. It can be seen that BUTD and \huapproachCaptioner without OCR inputs rarely mention text in the image except for common brand logos such as ``pepsi'' that are easy to recognize visually. On the other hand, the full \huapproachCaptioner approach learns to read text in the image and mention it in its generated captions.\footnote{More predictions from M4C-Captioner are presented in Supplemental (Fig.~\ref{fig:big_fig}).} Moreover, \huapproachCaptioner learns and recognizes relations between objects and is able to combine multiple OCR tokens into one complex description. For e.g., in Fig.~\ref{fig:m4c-pred}(d) the model uses a OCR token to correctly name a player who is blocking another player; in Fig.~\ref{fig:m4c-pred}(e) the model attempts to include and combine multiple tokens into a single message (``the \emph{track} is \emph{moved} in \emph{Kenosha}" instead of ``the word \emph{moved}, the word \emph{track}, and the word \emph{Kenosha} are on the sign"). In Fig.~\ref{fig:m4c-pred}(b) prediction is constructed fully from vocabulary, and even then the model counts similar objects and returns ``two pepsi bottles" instead of ``pepsi bottle and pepsi bottle". We also observe a large amount of mistakes in model predictions. Many mistakes are due to wrong scene understanding and object identification, which is a common problem in captioning algorithms. We also observe placing OCR tokens in the wrong object or semantic context in the caption (Fig.~\ref{fig:m4c-pred}(c, e)), incorrect repetition of an OCR token in a caption (Fig.~\ref{fig:m4c-pred}(a, e)), or insufficient use of them (Fig.~\ref{fig:m4c-pred}(f)) by the model. Some mistakes (as ``number \emph{3}" in Fig.~\ref{fig:m4c-pred}(d) are due to the errors of OCR detection algorithm and not the captioning model.
This points to many potential directions for future development on this challenging generative task, which requires visual and textual  understanding, requiring new model designs, conceptually different from previously existing captioning models.

\myparagraph{Transferring to COCO.}
We further qualitatively show that when integrated with other datasets such as COCO \cite{coco}, our dataset also enables text-based captioning on other datasets. 
In this setting, we experiment training \huapproachCaptioner (Table~\ref{tab:exp_textcaps_method_comparison}'s best) on both \datasetName dataset and COCO dataset together. We balance the number of samples seen by the model from both COCO and \datasetName during training, and apply the trained model on the COCO validation set.
COCO Captions mostly focus on visual objects but we show several examples where reading is necessary to describe the scene in Fig~\ref{fig:m4c-pred-coco}. When trained on the union of our dataset and COCO, the \huapproachCaptioner learns to generate captions containing text present in the images. On the other hand, the same model only describes visual objects without mentioning any text when trained on COCO alone. Quantitative results can be found in Supplemental (Sec.~\ref{sec:supp_coco}).

\section{Conclusion}
\emph{\TaskName} is a novel challenging task requiring models to read text in the image, recognize the image content, and comprehend both modalities jointly to generate a succinct image caption. To enable models to learn this ability and study this task in isolation, we collected \datasetName with \datasetNCaptionsInK captions. The captions include a mix of objects and/or visual scene descriptions in relation to OCR tokens copied or rephrased from the images.  In most cases, OCR tokens have to be copied and  related to the visual scene, but sometimes the OCR tokens  have to be understood, and sometimes  spatial or visual reasoning between text and objects in the image is required, as shown in our ablation study.
Our analysis also points out several challenges of this dataset: Different from other captioning datasets, nearly all our captions require  integration of OCR tokens, many are unseen (``zero-shot"). In contrast to TextVQA datasets, \datasetName requires generating long sentences and involves new technical challenges, including many switches between OCR and vocabulary tokens.

We find that current state-of-the-art image captioning models cannot read when trained on existing captioning dataset. 
However, when adapting the recent \huapproach VQA model to our task and training it on our \datasetName dataset, we are able to generate impressive captions on both \datasetName and COCO, which involve copying multiple OCR tokens and correctly integrating them in the captions. 
Our human evaluation confirms the result of the automatic metrics with very high correlation, and also shows that human captions are still significantly better than automatically generated ones, leaving room for many advances in future work, including better semantic understanding between image and text content, missing reasoning capabilities, and reading long text or single characters.

We hope our dataset with challenge server, available at \href{https://textvqa.org/textcaps}{textvqa.org/textcaps}, will encourage the community to design better image captioning models for this novel task and address its technical challenges, especially increasing their usefulness for assisting visually disabled people. 

\textbf{Acknowledgments.} We would like to thank Guan Pang and Mandy Toh for helping us with OCR ground-truth collection. We would also like to thank Devi Parikh for helpful discussions and insights.

\bibliographystyle{splncs04}
\bibliography{refs}

\newpage
\appendix
\counterwithin{figure}{section}
\counterwithin{table}{section}
\counterwithin{equation}{section}

\begin{center}
\Large 
\textbf{\datasetName: a Dataset for Image Captioning with Reading Comprehension}\\(Supplementary Material)
\par
\end{center}

\begin{figure*}[]
\centering

\begin{tabular}{ccc}
\subfloat[a framed picture with the year {[2012]} on it]{\includegraphics[width = 1.5in]{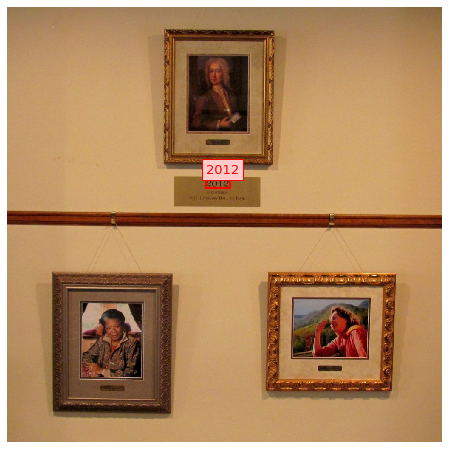}} &
\subfloat[a pair of {[merrell]} brand products are on a table]{\includegraphics[width = 1.5in]{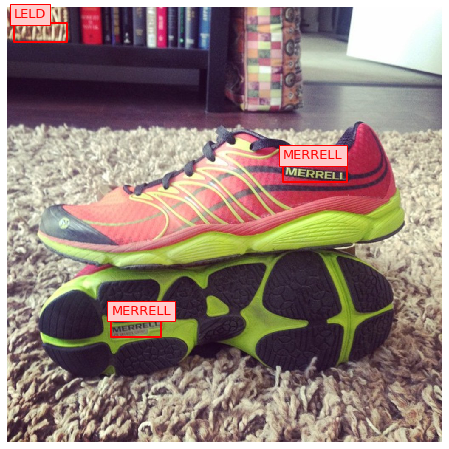}} &
\subfloat[a bottle of {[deluse]} sits on a table next to a small small small plastic bottle]{\includegraphics[width = 1.15in]{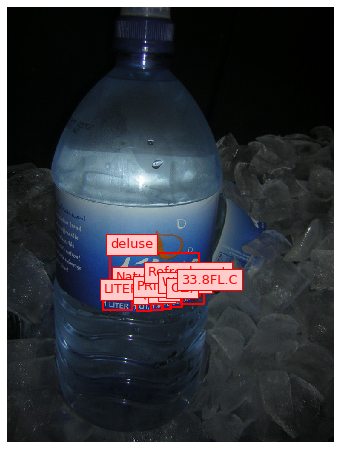}}\\
\subfloat[a woman is looking at a screen that says {[£20]} on it]{\includegraphics[width = 1.5in]{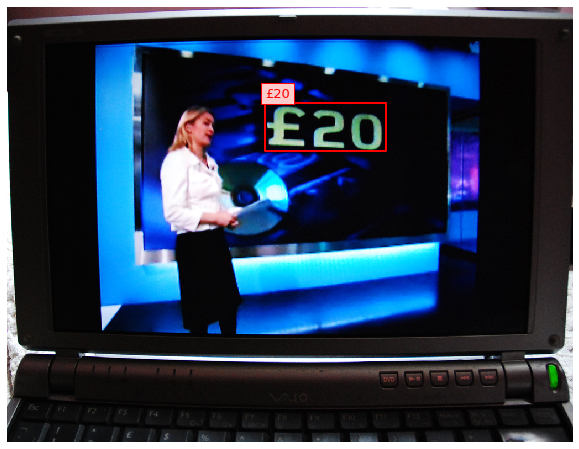}} &
\subfloat[a trash can with a sticker on it that says {[seniors]} {[only]}]{\includegraphics[width = 1.5in]{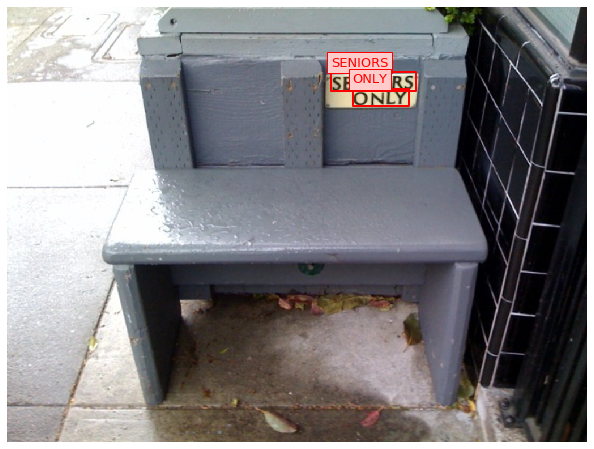}} &
\subfloat[a sign that says ' {[cash]} {[custor]} {[pay]} {[firs]} ' on it]{\includegraphics[width = 1.5in]{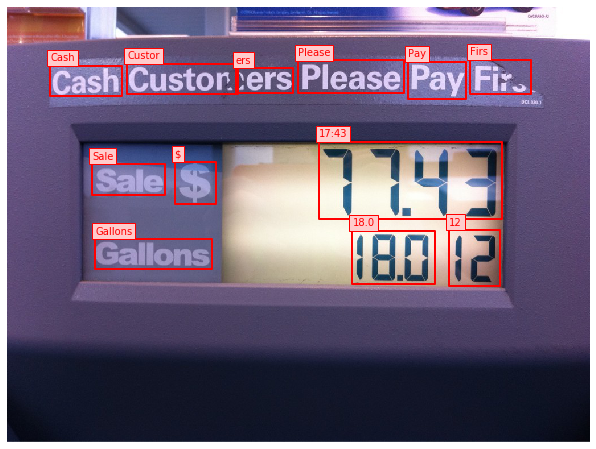}}\\
\subfloat[a plane with the number {[202]} on the side of it]{\includegraphics[width = 1.5in]{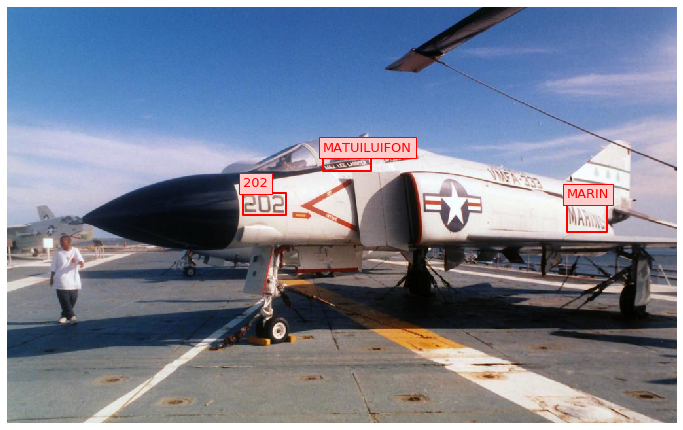}} &
\subfloat[a red telephone booth with a red telephone booth]{\includegraphics[width = 1.5in]{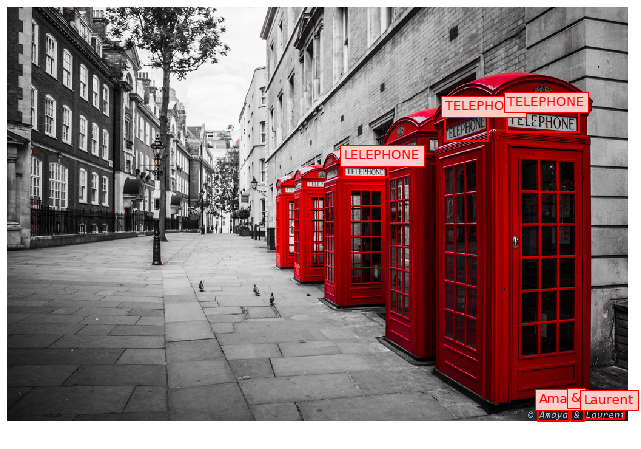}} &
\subfloat[a restaurant with a red sign that says {[bar]} {[bar]}]{\includegraphics[width = 1.5in]{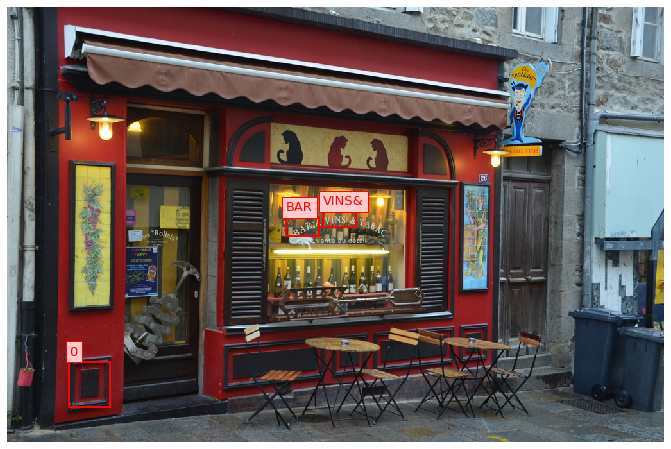}}\\
\subfloat[a man is holding a box that says ' i ' m ' on it]{\includegraphics[width = 1.5in]{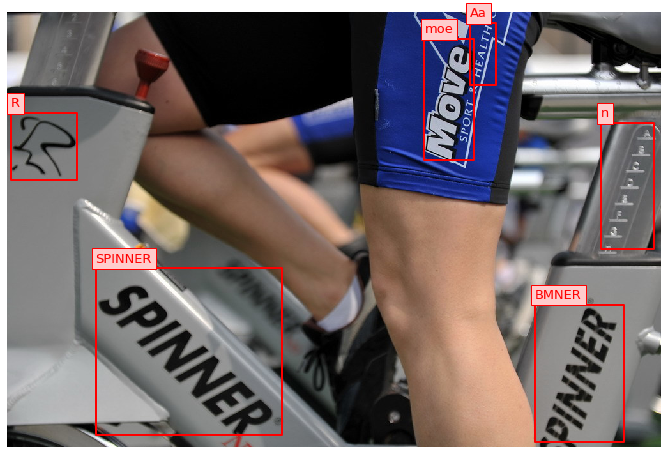}} &
\subfloat[a black box with the word {[bizhub]} on it]{\includegraphics[width = 1.5in]{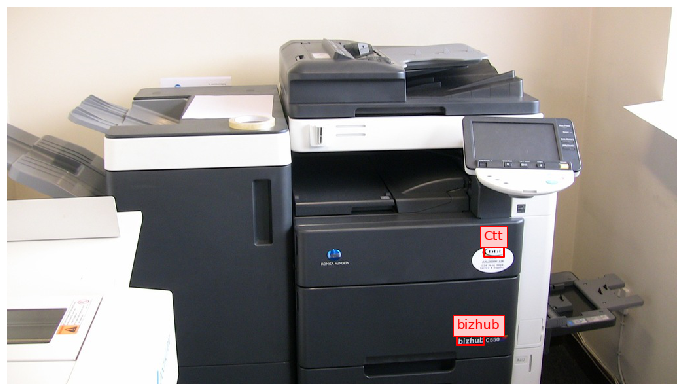}} &
\subfloat[a bottle of wine with the word {[chenet]} on the label]{\includegraphics[width = 1.15in]{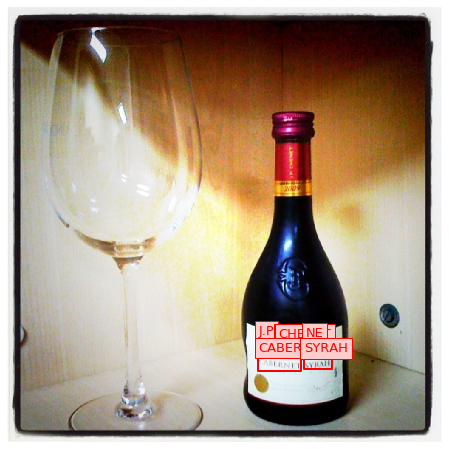}}
\end{tabular}
\caption{\textbf{Additional examples of M4C-Captioner predictions on the test set.} Square brackets denote tokens which model selected from OCR tokens while others are from vocabulary.}
\label{fig:big_fig}
\end{figure*}

We include the following material in our supplemental material:
\begin{itemize}
\item[\textbf{Section \ref{sec:gt-ocr}.}] Analysis of the influence of GT-OCR on \huapproachCaptioner model.
\item[\textbf{Section \ref{sec:rosetta_ocr_performance}.}] Precision and recall of Rosetta OCR tokens on TextCaps.
\item[\textbf{Section \ref{sec:supp_coco}.}] We show the evaluation on the COCO dataset: Our qualitative (Figure \ref{fig:supp_coco}) and quantitative (Table \ref{tab:supp_coco}) results show that COCO references captions rarely involve reading comprehension, indicating that COCO captions are not a good dataset for training or evaluating this task.
\item[\textbf{Section \ref{sec:wordclouds}.}] Qualitative illustration of frequent words in TextCaps images and captions.
\item[\textbf{Section \ref{sec:test-val}.}] Comparison of TextCaps Test and Validation sets.
\item[\textbf{Section \ref{sec:user-interfache}.}] Data collection User Interface.
\item[\textbf{Figure \ref{fig:big_fig}.}] Additional examples of M4C-Captioner predictions.
\end{itemize}

\section{Analysis of the influence of GT-OCR on \huapproachCaptioner model}
\label{sec:gt-ocr}

In this section, we provide additional analysis on ground-truth OCRs (GT-OCRs). So far we collected GT-OCR annotations for around 96\% on the training, 96\% on the validation, and 92\% on the test set, we excluded OCR annotation of non-Latin/non-English characters. 
In Table 1 in the main paper, ``M4C-Captioner (w/ GT OCRs)'' (line 14 and 18) is evaluated on all TextCaps validation and test set images respectively, where an empty OCR list is used as inputs to the model on those images without GT-OCR annotations. Here, we also specifically compare the methods on the subsets of TextCaps validation and test sets, excluding those images with empty GT-OCR annotations. The results are shown in Table \ref{tab:m4c_gt_ocr}, where ground truth OCR tokens improve the quality of generated predictions significantly. 

\begin{table}[h]
\setlength{\tabcolsep}{3pt}
\scriptsize
\caption{\textbf{Effect of using GT-OCR on performance of \huapproachCaptioner on \datasetName dataset.} Evaluated on a subsets of images with GT-OCR annotations (96\% for the validation set, 92\% for the test set).}
\begin{center}
\begin{tabular}{cp{7.0cm}cccccc}
\cmidrule[\heavyrulewidth]{1-7}
&  & \multicolumn{5}{c}{\datasetName validation set metrics}  \\
\cmidrule{3-7} 
\# & Method & B-4 & M & R & S & C \\
\cmidrule{1-7}
1 & \huapproachCaptioner 
& 23.0 & 21.9 & 46.1 & 15.4 & 88.7 \\ 
2 & \huapproachCaptioner (evaluated w/ GT OCRs)
& 24.4 & 22.8 & 47.0 & 16.2 & 99.4 \\
3 & \huapproachCaptioner (trained and evaluated w/ GT OCRs)
& 26.3 & 23.3 & 48.0 & 16.4 & 107.2 \\ 
\cmidrule{1-7}
&  & \multicolumn{5}{c}{\datasetName test set metrics}  \\
\cmidrule{3-7} 
\# & Method & B-4 & M & R & S & C \\
\cmidrule{1-7}
4 & \huapproachCaptioner 
& 19.0 & 19.7 & 43.2 & 12.7 & 80.7 \\ 
5 & \huapproachCaptioner (evaluated w/ GT OCRs)
& 20.4 & 20.8 & 44.5 & 13.7 & 95.1 \\
6 & \huapproachCaptioner (trained and evaluated w/ GT OCRs)
& 22.2 & 21.6 & 45.8 & 14.0 & 103.5 \\
\midrule
7 & Human
& 24.4 & 26.1 & 46.9 & 18.8 & 125.1 \\ 
\bottomrule
\end{tabular} \\
B-4: BLEU-4; M: METEOR; R: ROUGE\_L; S: SPICE; C: CIDEr
\end{center}
\label{tab:m4c_gt_ocr}
\end{table}

Given that we would like to have a self-contained captioning model at test time without human in the loop, we also investigate whether ground-truth OCRs can be helpful if they are only available at training time but not at test time. In Table~\ref{tab:exp_training_on_gt_ocr_vs_rosetta}, we compare two \huapproachCaptioner models trained with automatic (Rosetta) OCRs and GT-OCRs respectively and tested both with automatic (Rosetta) OCRs. It can be seen the results are very close between the two cases, suggesting that direct training on GT-OCRs does not improve the performance out-of-the-box.

\begin{table}[bth]
\setlength{\tabcolsep}{3pt}
\scriptsize
\caption{\textbf{Comparison between using automatic (Rosetta) OCRs and GT-OCRs during training.} We experiment with \huapproachCaptioner trained on our \datasetName dataset, using either automatic (Rosetta) OCRs or GT-OCRs during training. At test time, only automatic OCRs are provided, and no GT-OCRs are used in either case. Line 1 and 3 in this table are the same as line 13 and 17 in Table 1 of the main paper.}
\begin{center}
\begin{tabular}{cp{5.3cm}ccccc}
\cmidrule[\heavyrulewidth]{1-7}
& & \multicolumn{5}{c}{\datasetName validation set metrics}  \\
\cmidrule{3-7} 
\# & Method 
& B-4 & M & R & S & C \\
\cmidrule{1-7}
1 & \huapproachCaptioner
& 23.3 & \textbf{22.0} & \textbf{46.2} & \textbf{15.6} & \textbf{89.6} \\ 
2 & \huapproachCaptioner (trained w/ GT-OCRs) & \textbf{23.9} & 21.9 & 46.1 & 15.1 & 88.6 \\
\cmidrule[\heavyrulewidth]{1-7} 
& & \multicolumn{5}{c}{\datasetName test set metrics} \\
\cmidrule{3-7} 
\# & Method 
& B-4 & M & R & S & C \\
\cmidrule{1-7}
3 & \huapproachCaptioner 
& 18.9 & \textbf{19.8} & 43.2 & \textbf{12.8} & 81.0 \\
4 & \huapproachCaptioner (trained w/ GT-OCRs) & \textbf{19.7} & \textbf{19.8} & \textbf{43.6} & 12.4 & \textbf{81.9} \\
\bottomrule
\end{tabular} \\
B-4: BLEU-4; M: METEOR; R: ROUGE\_L; S: SPICE; C: CIDEr
\end{center}
\label{tab:exp_training_on_gt_ocr_vs_rosetta}
\end{table}

The analysis of predictions from M4C-Captioner model with automatically extracted and ground-truth OCR tokens (trained and evaluated) shows that vocabulary size of all tokens used in predictions and OCR tokens in particular does not change significantly (Table \ref{tab:gt-ocr-preds}). Although, the quality of OCR tokens used increase, as indicated by the precision metric. Precision is calculated as average ratio of OCR tokens predicted by the model which match OCR tokens used by annotators from total number of OCR predicted in each sentence.

\begin{table}[]
\setlength{\tabcolsep}{3pt}
\scriptsize
\caption{\textbf{Statistics of predicted sentences with automatic OCR compared to GT-OCR}}
\vspace{-1em}
\begin{center}
\begin{tabular}{lccccc}
\toprule
 & \multicolumn{3}{c}{vocab size} & OCR  \\
Method & Total  & OCR  & VOCAB  & token precision \\
\cmidrule(lr){1-1}\cmidrule(lr){2-4}\cmidrule(lr){5-5}
\huapproachCaptioner & 3287 &  2957  & 545 & 0.62 \\
\huapproachCaptioner (trained\&evaluated w/ GT OCRs) & 3391 &  3106 &  491 & 0.78 \\
\bottomrule
\end{tabular} 
\end{center}
\label{tab:gt-ocr-preds}
\end{table}

\section{Rosetta OCR performance analysis}
\label{sec:rosetta_ocr_performance}

In our experiments, we use Rosetta \cite{rosetta} to extract OCR tokens from an image. The English-only version of Rosetta is used, which is referred to as \textbf{Rosetta-en} in \cite{hu}. To measure the performance of the Rosetta OCR system on TextCaps, we evaluated the precision and recall of OCR tokens against the human-annotated text (ground-truth OCRs) over the validation and test set images, following the ICDAR-13 evaluation protocol for end-to-end text recognition \cite{karatzas2013icdar}. On the validation set images, the Rosetta OCR tokens have a precision of 56.50, a recall of 37.15, and an F-1 score of 44.83. On the test set images, the Rosetta OCR tokens have a precision of 53.60, a recall of 36.92, and an F-1 score of 43.72.

\section{Automatic evaluation on COCO captioning}
\label{sec:supp_coco}

\begin{table}[b!]
\setlength{\tabcolsep}{3pt}
\scriptsize
\caption{Automatic evaluation metrics of the \huapproachCaptioner model on the COCO captioning validation set (Karpathy split). Here, training on \datasetName leads to lower metrics on COCO. This is mainly because the human captions in the COCO dataset do not involve reading comprehension in addition to the domain shift between the two datasets. See Sec.~\ref{sec:supp_coco} and Figure~\ref{fig:supp_coco} for details.}
\begin{center}
\begin{tabular}{cp{4cm}cccccc}
\toprule
\# & Method & Trained on & B-4 & M & R & S & C \\
\midrule
1 & \huapproachCaptioner & COCO & \textbf{34.3} & \textbf{27.5} & \textbf{56.2} & \textbf{20.6} & \textbf{112.2} \\
2 & \huapproachCaptioner & COCO+\datasetName & 27.1 & 24.1 & 51.6 & 17.4 & 87.5 \\
\bottomrule
\end{tabular}
~\\
\vspace{0.2em}
\small{B: BLEU-4; M: METEOR; R: ROUGE\_L; S: SPICE; C: CIDEr} \\
\end{center}
\vspace{-0em}
\label{tab:supp_coco}
\end{table}

\begin{figure}[]
\centering

\includegraphics[width=\linewidth]{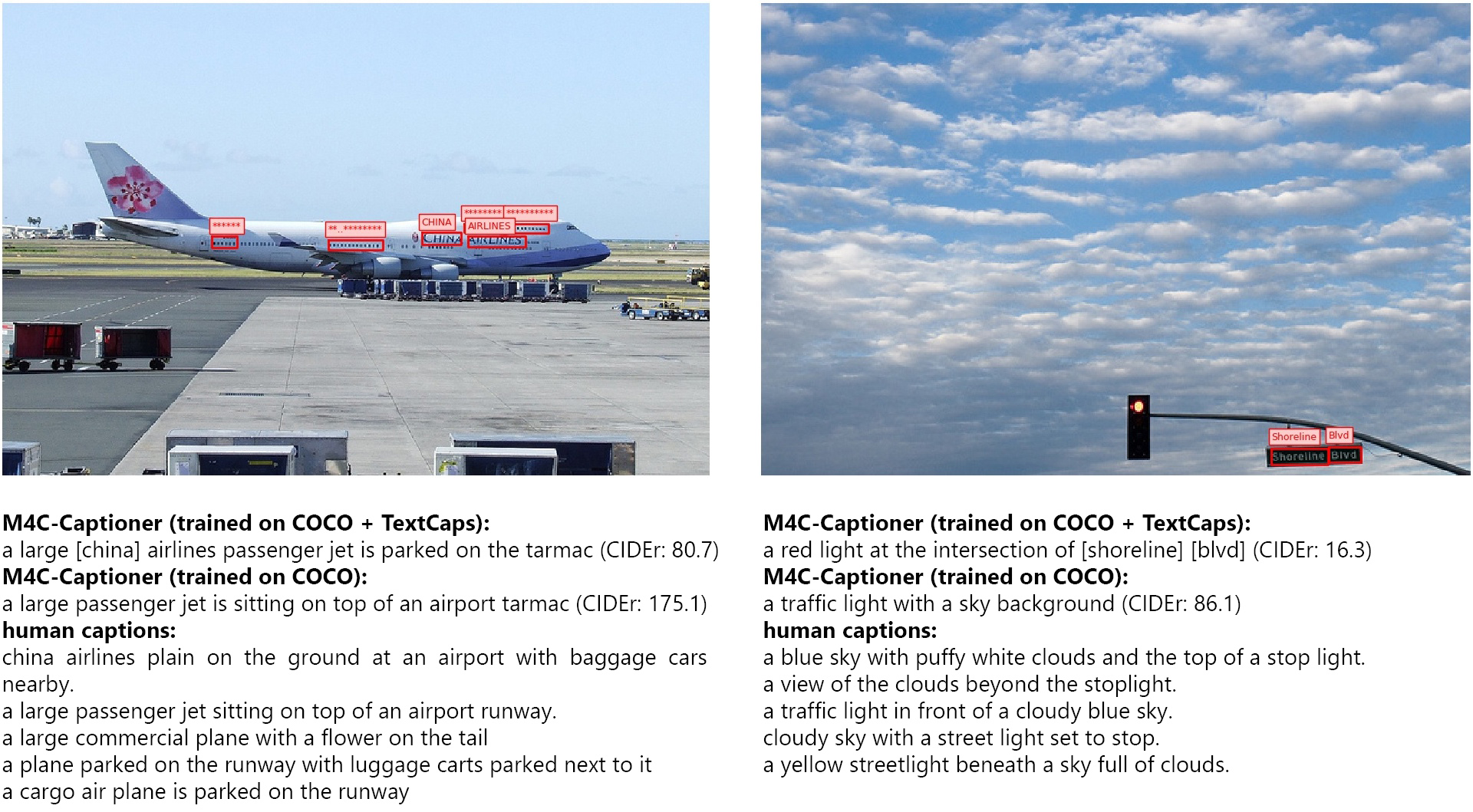}

\caption{Predicted and human captions on COCO validation set (Karpathy split), where words copied from OCR tokens are taken in square brackets. As human captions in COCO rarely describe text in the image, generated captions that mention text often have lower CIDEr scores.}
\label{fig:supp_coco}
\end{figure}

Table~\ref{tab:supp_coco} shows the automatic evaluation metrics of the \huapproachCaptioner model on the COCO dataset. Here, the model trained on COCO + \datasetName\footnote{When training on COCO + \datasetName in this setting, we sample TextCaps captions more frequently than COCO captions to encourage learning text reading.} has lower evaluation scores than the same model trained only on COCO. We also experiment with different sampling ratios between COCO captions and TextCaps captions, and observe that higher TextCaps ratio (sampling TextCaps captions more frequently) leads to better qualitative results where more OCR tokens are described in the generated captions, but worse CIDEr scores on the COCO validation set. We inspect and find that this is mainly because the human captions in the COCO dataset rarely involve reading comprehension. For example, in Figure~\ref{fig:supp_coco}, we see that the predicted captions from \huapproachCaptioner trained on COCO + \datasetName has noticeably lower CIDEr scores, although it learns to read and copy relevant text from the image.

\begin{figure}[b!]
\begin{center}
\includegraphics[width=\linewidth]{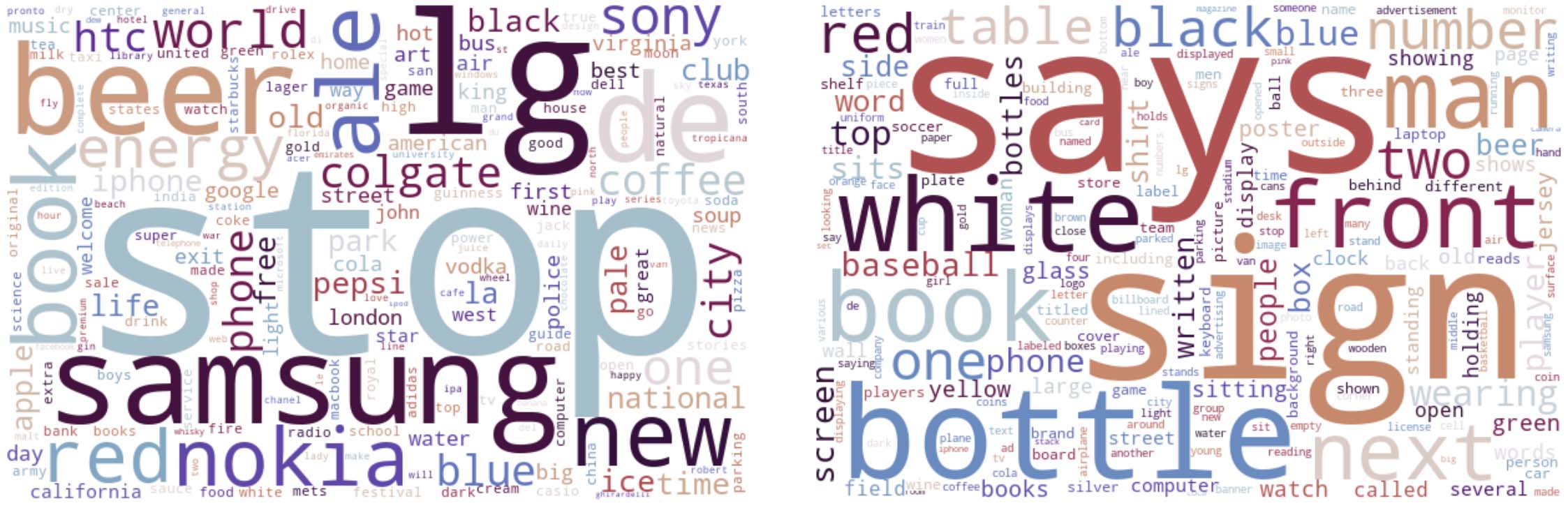}
\end{center}
\vspace{\figcapmargin}
\caption{\textbf{Wordcloud visualizations} of most frequent OCR tokens (left) and all words (right) in \datasetName captions.}
\label{fig:wordcloud}
\end{figure}

\section{Illustration of most frequent words in TextCaps}
\label{sec:wordclouds}
We visualize word clouds for the text tokens in \datasetName captions in Fig.~\ref{fig:wordcloud}. In the left word cloud, it can be seen that  OCR tokens  copied from the image to the caption with high frequency  mainly consist of brand names and other words which can be found on the products and their labels (\myquote{samsung}, \myquote{nokia}, \myquote{colgate}, \myquote{ale}). In the right word cloud, when all the words  are taken into account, we can observe great use of words like \myquote{sign}, \myquote{says}, \myquote{written} which annotators used to incorporate text tokens into their captions.

\section{Comparison of TextCaps Test and Validation sets}
\label{sec:test-val}
We notice a difference in performance on our validation and test set in Table~1 of the main paper, specifically, the performance on the validation set is always higher. In this section, we discuss the similarities and differences of the two sets to understand the performance difference.

\begin{figure}[]
\begin{center}
\includegraphics[width=\linewidth]{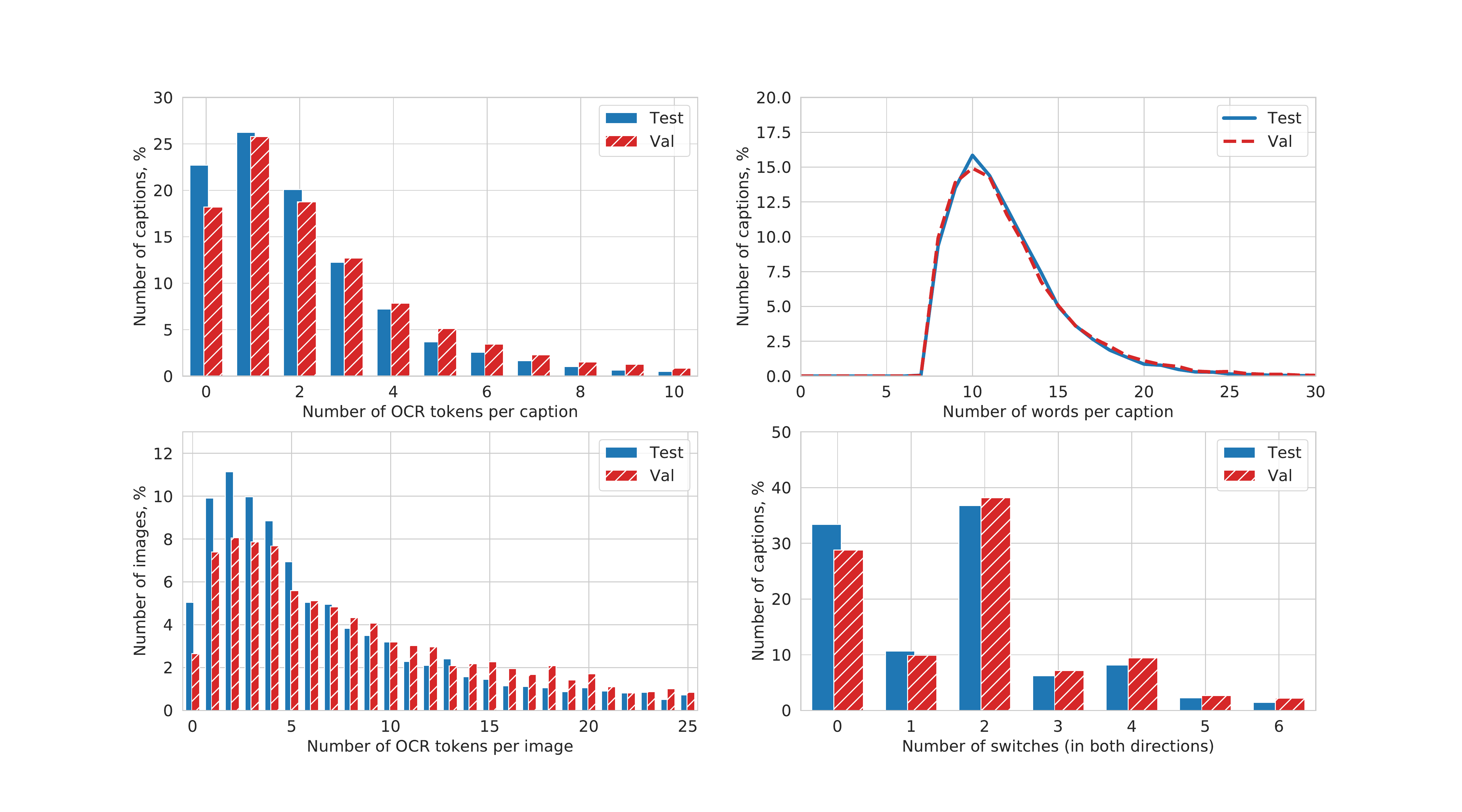}
\end{center}
\vspace{\figcapmargin}
\caption{\textbf{Statistics of \datasetName Test and Validation sets.}}
\label{fig:valtest}
\end{figure}

First, the images for TextCaps `training' and `validation' sets are from the  OpenImages \cite{openimages} `training' set, while \datasetName `test' images are from the OpenImages `test' set. We observe that the image-labels of OpenImages training and test sets have slightly different distributions and categories \cite{openimages-figs}. 
As our training and validation set are both from the same image distribution (as we follow TextVQA's split \cite{textvqa}), it is likely that models trained on the training set better fit the validation set than the test set.
This is partially confirmed by evaluating a model trained only on COCO captions on TextCaps validation and test set. Here, the performance difference is smaller, for example for the BUTD model the CIDEr score drops by 4\% (relative) from validation to test set when trained on COCO, but 20\% when trained on TextCaps.

Second, although  the captions for training, validation, and test were collected jointly, the  different image distributions might affect the captions. For this, we further compare their statistics. In particular, we  observe from Figure \ref{fig:valtest} that both images and captions of the validation set have a larger number of OCR tokens on average (note that all these statistics are based on automatically extracted OCR tokens). This also causes a larger number of switches between OCR and vocabulary required in the validation set. On the other hand, in the test set we observe more captions without any copied OCR tokens, which could suggest more paraphrasing, reasoning, and re-formulation of the OCR tokens in this set. 
The distribution of captions' length is almost the same for both sets. 

Third, we evaluate and compare the automatic metrics on human-written captions between the test and validation sets. Since there are only 5 human captions (instead of 6) collected on the validation set, we perform a similar leave-one-out evaluation as mentioned in Sec. 4.1 but using only 5 human captions per image (evaluating 1 human caption over the remaining 4 and averaging over the 5 runs). The results are shown in Table \ref{tab:human_on_val_and_test}, where the BLEU-4, METEOR, ROUGE\_L, and CIDEr metrics are higher on the test set than on the validation set. This is a bit surprising, but also indicates, that there is slight domain shift between validation and test set, which humans are not affected by, rather than that the test set is more difficult in itself as the models' performance drop might suggest.

\begin{table}[h]
\setlength{\tabcolsep}{3pt}
\scriptsize
\caption{\textbf{Comparison of automatic metrics on human captions between \datasetName test and validation set}, using 5 human captions per image (evaluating 1 human caption over the remaining 4 and averaging over the 5 runs).}
\vspace{-1em}
\begin{center}
\begin{tabular}{cp{6.3cm}cccccc}
\cmidrule[\heavyrulewidth]{1-8}
\cmidrule{4-8} 
\# & Method &  
& B-4 & M & R & S & C \\
\cmidrule{1-8}
1 & Human captions on the TextCaps validation set &  
& 22.1 & 24.8 & 44.6 & 20.3 & 118.0 \\
2 & Human captions on the TextCaps test set &  
& 22.6 & 25.4 & 45.5 & 20.3 & 127.9 \\
\bottomrule
\end{tabular} \\
\vspace{0.2em}
\small{B-4: BLEU-4; M: METEOR; R: ROUGE\_L; S: SPICE; C: CIDEr}
\end{center}
\vspace{-2em}
\label{tab:human_on_val_and_test}
\end{table}

\section{Data collection User Interface}
\label{sec:user-interfache}
\begin{figure*}[]
\begin{center}
\includegraphics[width=\linewidth]{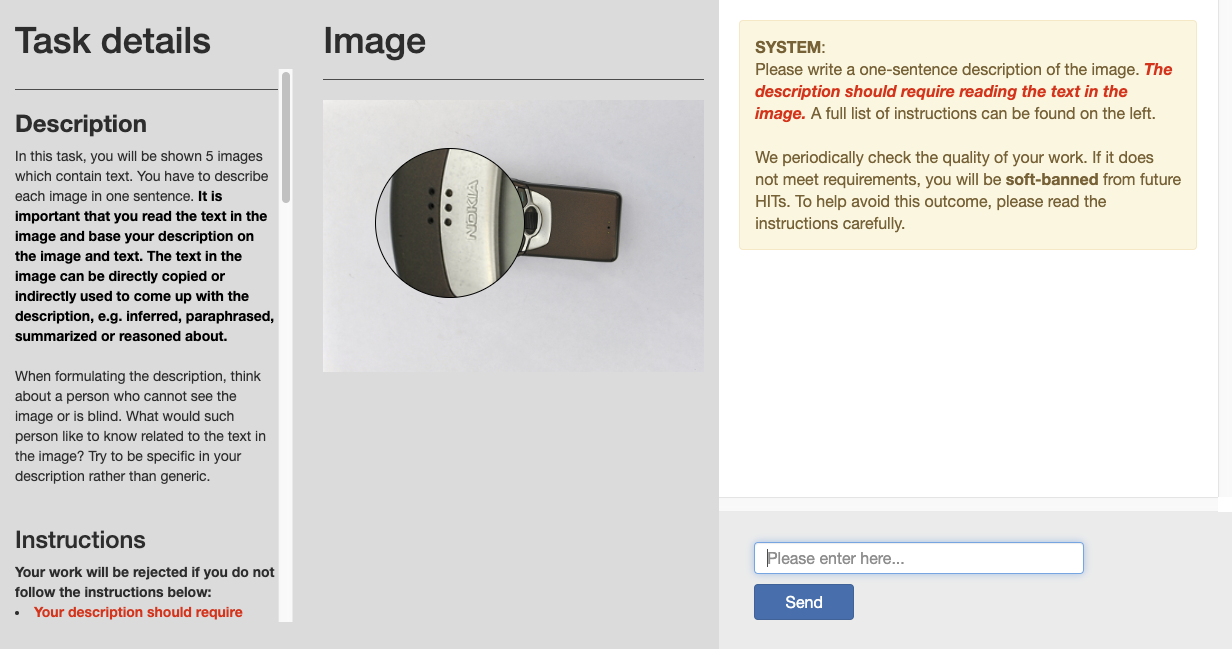}
\end{center}
\vspace{\figcapmargin}
\caption{\textbf{The main interface window for the annotation stage of the data collection.} Detailed instructions are shown in the next Figure. The circle is an interactive zoom tool which users can move with the mouse cursor like a magnifier lens.}
\label{fig:write2}
\end{figure*}

\begin{figure*}[]
\begin{center}
\includegraphics[width=0.9\linewidth]{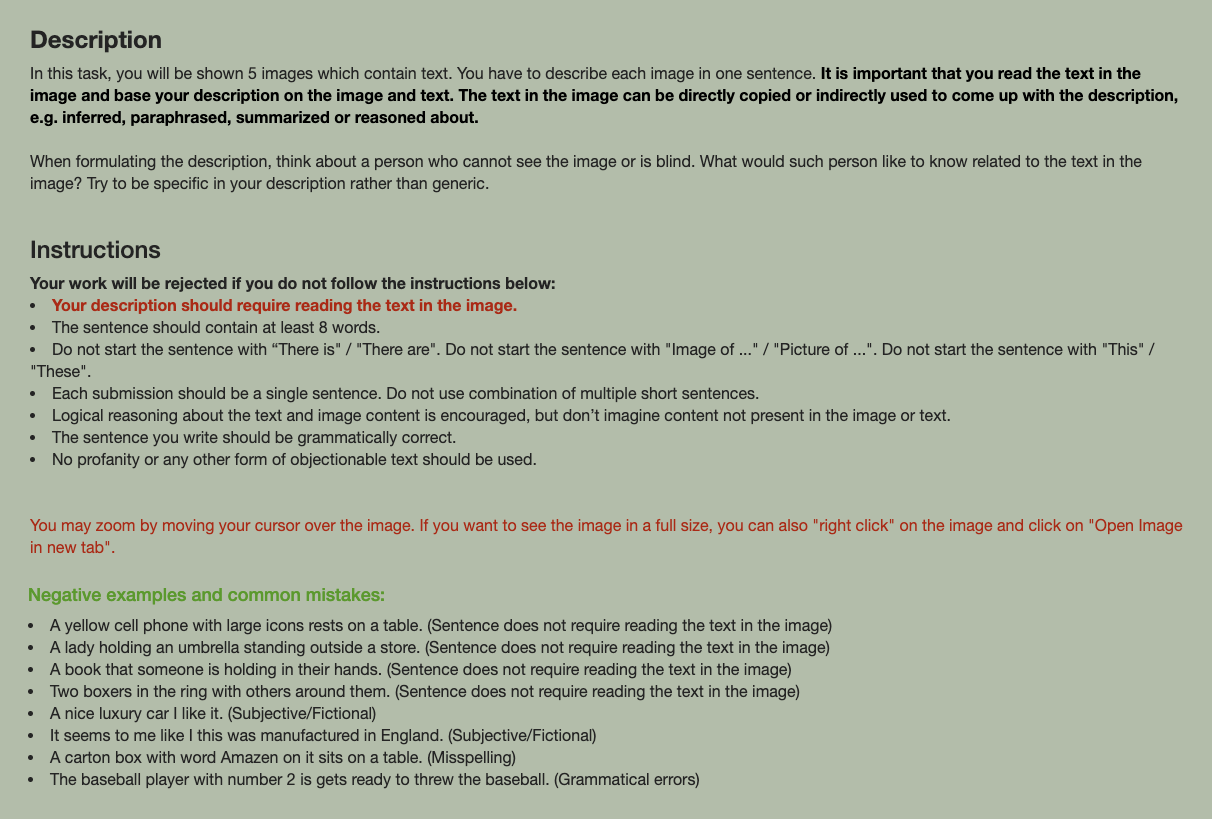}
\end{center}
\vspace{\figcapmargin}
\caption{\textbf{Instructions for the annotation stage of data collection.} First time users saw the instructions before starting the task, after which they could find it on the left panel of our main task interface.}
\label{fig:write1}
\end{figure*}

\begin{figure*}[]
\begin{center}
\includegraphics[width=\linewidth]{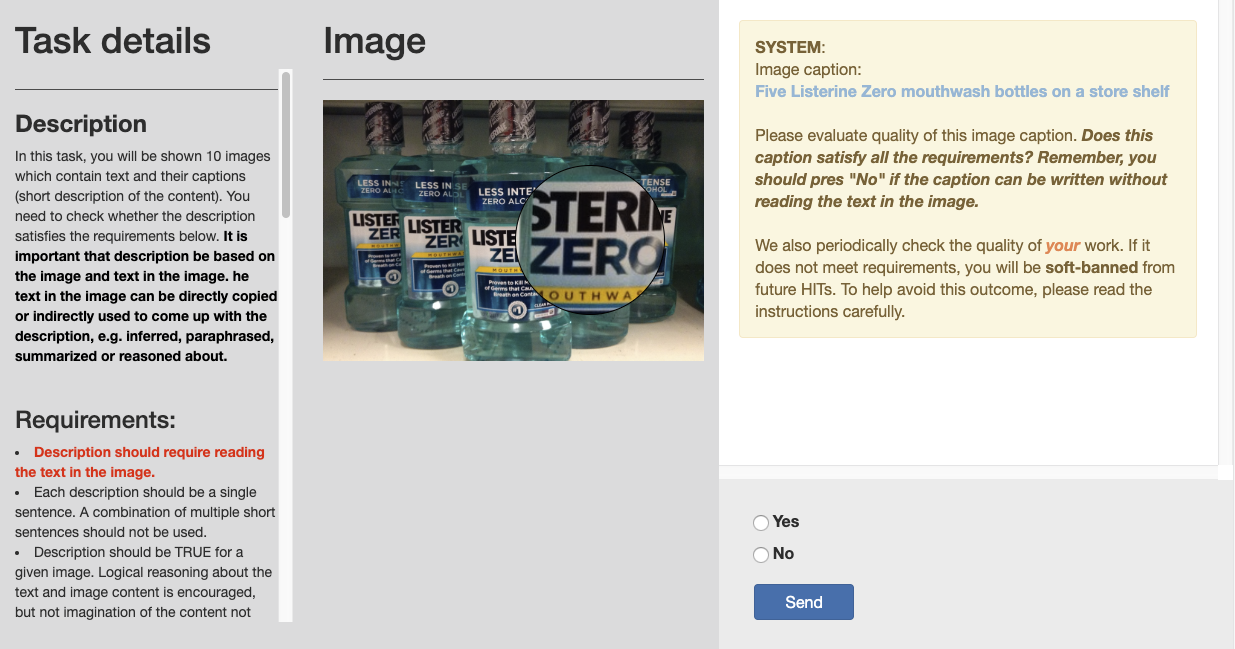}
\end{center}
\vspace{\figcapmargin}
\caption{\textbf{The main interface window for the evaluation stage of the data collection.} Detailed instructions are shown in the next Figure. The circle is an interactive zoom tool which users can move with the mouse cursor like a magnifier lens.}
\label{fig:eval2}
\end{figure*}

\begin{figure*}[]
\begin{center}
\includegraphics[width=\linewidth]{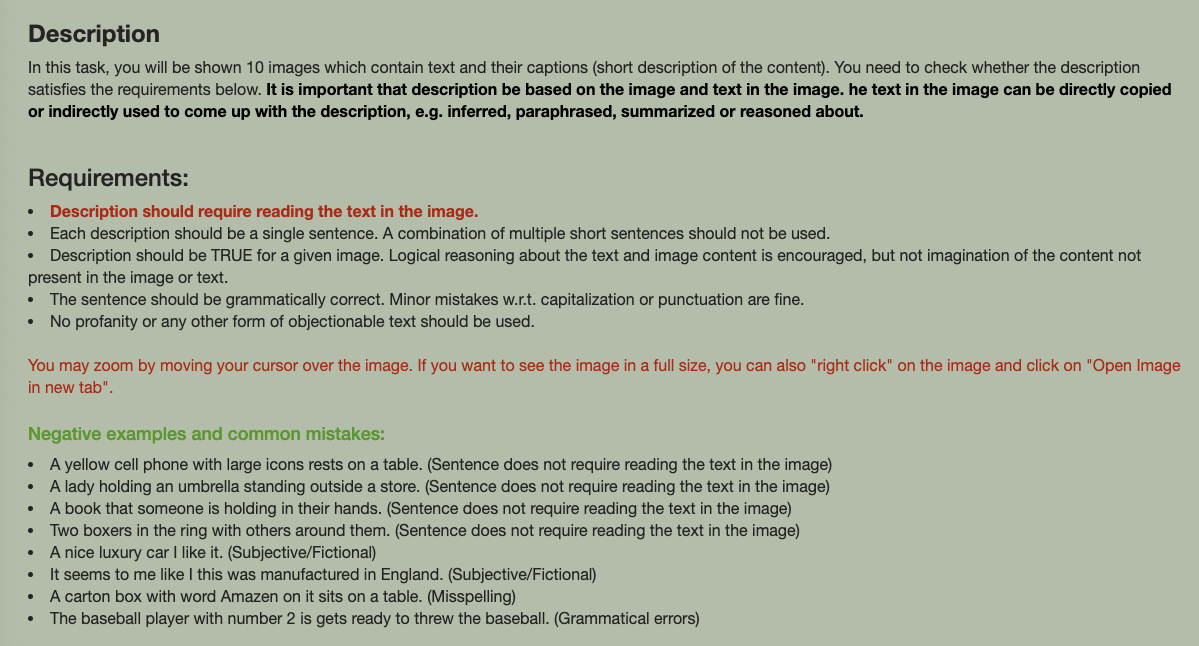}
\end{center}
\vspace{\figcapmargin}
\caption{\textbf{Instructions for the evaluation stage of data collection.} First time users saw the instructions before starting the task, after which they could find it on the left panel of our main task interface.}
\label{fig:eval1}
\end{figure*}

\datasetName was collected using the interfaces presented in Figures~\ref{fig:write2}-\ref{fig:eval1}. Before starting, users were presented with a list of detailed instructions (Fig.~\ref{fig:write2} and Fig.~\ref{fig:eval2} for annotation and evaluation, respectively). The main interface window includes a panel with the same list of instructions on the left, a panel with an image in the center, and short instruction followed by the answer field on the right (Fig.~\ref{fig:write1},~\ref{fig:eval1}). It is worth noting that in the case of small or hardly-readable text users had options either to open the image in a full size in a new window or to use interactive magnifier lens with 3x zoom. Users annotated images in  mini-batches of 5, and evaluated captions in mini-batches of 10. There was no restriction by time (except of very extreme limit of 15 minutes per mini-batch).

\end{document}